# Attribution assignment for deep-generative sequence models enables interpretability analysis using positive-only data

Robert Frank[1,c], Michael Widrich[2], Rahmad Akbar[3], Günter Klambauer[2], Geir Kjetil Sandve[4], Philippe A. Robert[5,c], Victor Greiff[1,6,c]

[1] Department of Immunology and Oslo University Hospital, University of Oslo, Oslo, Norway
[2] Institute for Machine Learning, Johannes Kepler University Linz, Linz, Austria
[3] Novo Nordisk A/S, Denmark
[4] Department of Informatics, University of Oslo, Oslo, Norway
[5] University Hospital Basel, Basel, Switzerland
[6] Imprint Labs, LLC. New York, NY, USA
[C] To whom correspondence should be addressed

# Abstract

Generative machine learning models offer a powerful framework for therapeutic design by efficiently exploring large spaces of biological sequences enriched for desirable properties. Unlike supervised learning methods, which require both positive and negative labeled data, generative models such as LSTMs can be trained solely on positively labeled sequences, for example, high-affinity antibodies. This is particularly advantageous in biological settings where negative data are scarce, unreliable, or biologically ill-defined. However, the lack of attribution methods for generative models has hindered the ability to extract interpretable biological insights from such models. To address this gap, we developed Generative Attribution Metric Analysis (GAMA), an attribution method for autoregressive generative models based on Integrated Gradients. We assessed GAMA using synthetic datasets with known ground truths to characterize its statistical behavior and validate its ability to recover biologically relevant features. We further demonstrated the utility of GAMA by applying it to experimental antibody-antigen binding data. GAMA enables model interpretability and the validation of generative sequence design strategies without the need for negative training data.

# Introduction

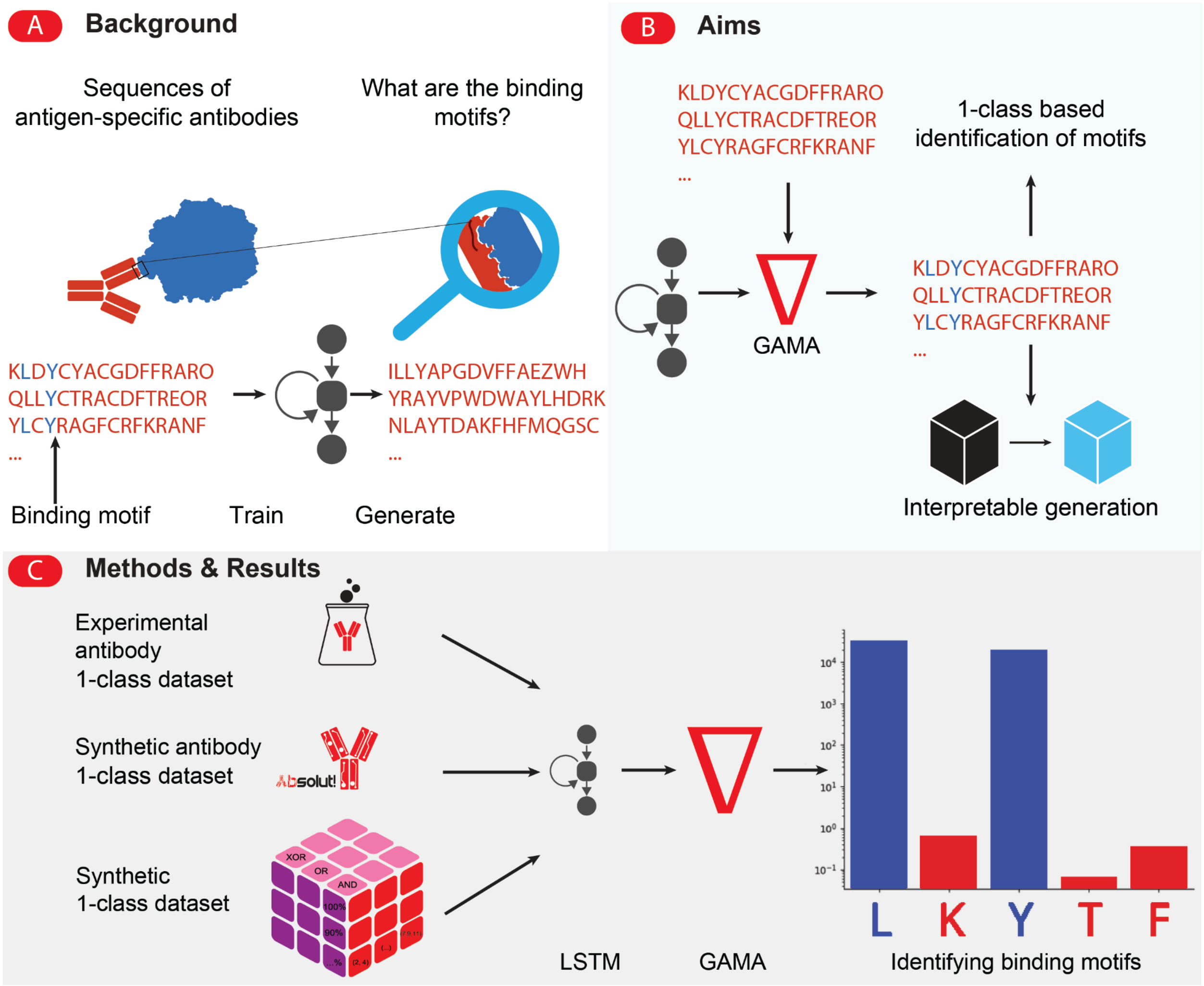

**Figure 1. GAMA is an attribution assignment method for generative sequence models.** (A) Deep generative sequence models enable the design of novel biological sequences from only positively labeled datasets, such as antibody CDR sequences. However, deep generative models are black box models and it is unclear what patterns they learn. We investigate how to interrogate the learned distribution of generative sequence models and how generative sequence models can be used to understand the training distribution. (B) We developed a method called "Generative Attribution Metric Analysis" (GAMA) to identify learned motifs in generative models and make them more interpretable. Generative models only require 1-class data that consist solely of positive instances. For example, antibody binding datasets have only instances of antigen-binding (positive) sequences available but it is difficult to experimentally acquire sequences that explicitly do not bind the antigen. (C) We evaluate GAMA on 270 synthetic sequence datasets, four synthetic antibody sequence datasets, and one experimental antibody sequence dataset to demonstrate retrieval efficiency and interpretability.

Deep generative neural networks are a class of models capable of learning complex data distributions and can enable the generation of novel instances of that distribution. They do not require labels and can be trained on "positive-only samples" (hereafter referred to "1-class samples"). Some of the most common approaches used in generative modeling are Long Short-Term Memory networks (LSTM) [1], state space models[2], Generative Adversarial Networks (GAN) [3], Variational AutoEncoder (VAE) [4], diffusion models [5], and transformer networks [6].

Several protein design approaches have been successful using deep (generative) models [7–19]. Generative models have also been used to design DNA and RNA sequences, e.g., to design novel promoter sequences or RNA sequences with desired folding properties [20–23]. Previous reports have demonstrated that monoclonal antibodies can be synthetically generated using generative models and have a distribution of binding affinity and various biophysical parameters that is similar to the training data distribution [7,8,24–30].

The interpretability of deep learning methods is inherently challenging, as they process information in hierarchically stacked, highly nonlinear representations. Various attribution assignment methods have been developed to address this issue, aiming to elucidate the contribution of input features to the output decision [31–33]. These methods are instrumental in diagnosing individual classification errors and assessing the confidence level of model reliability [34]. Among these techniques, Integrated Gradients (IG) stands out as a particularly effective approach [35]. IG emphasizes the significance of input features in the prediction process and has been successfully employed in various supervised sequence prediction tasks. For instance, Senior and colleagues demonstrated IG's utility in elucidating the influence of feature location on protein folding prediction from sequences [36,37]. Similarly, Preuer and colleagues showcased IG's role in molecule design for drug discovery [38], and Ishida and colleagues utilized IG to identify key atoms in molecules for reaction prediction [39].

The use of IG in supervised LSTM models is well-documented across multiple studies [40,41], including its application to DNA sequences [42] and RNA sequences [43]. IG has also been implemented in the analysis of antibody sequences: our recent work revealed that IG, when employed in binary classification methods, uncovers structural patterns relevant to antibody binding [44,45]. Furthermore, Widrich and colleagues identified pertinent patterns in amino acid sequences within immune repertoire datasets using IG [46]. A detailed guide on integrating the attribution assignment method "Layer-wise Relevance Propagation" into LSTM models was provided by Arras and colleagues [47].

Extensive research has been performed on rendering supervised methods interpretable. However,there is only limited research available that has focused on the interpretability of generative sequence models [48,49],especially generative sequence models for 1-class data.[50] There are several papers on developing attribution assignment methods for transformers and large language models [51–54]. There exists some work on interpreting generative (sequence) models with a latent space [55–57]. Approaches that tackle attribution assignment methods for 1-class generative models are rare [57,58]. A survey paper from Tritscher et al. is available [59]. To our knowledge, there exists no prior work which uses gradient-based methods on LSTMs for attempting to identify essential patterns that are relevant for LSTM-based sequence generation.

Interpretable methods would enable addressing the negative label bias problem, which primarily occurs in immunological datasets but poses a challenge in many supervised multiclass settings where the definition of negative data is not unambiguous [44,45,60–62]. Specifically, supervised methods for immune receptor classification tasks require at least two labels in the training data, where one class usually consists of those receptor sequences that have been validated to bind the target antigen. The other, negative class, is challenging to define as it may be defined as non-binders or, for example, binders to another target antigen [45]. Indeed, we and others have previously shown that the training data composition strongly impacts the classification of the positive class by ML methods and best practices for defining negative datasets are still controversial [44,45,60–62]. Applying attribution assignment methods to solely 1-class data (i.e., antigen-specific binders) using generative models would reduce the requirements for training data size and complexity, enabling the generation of both novel binders and an understanding of the underlying binding patterns. Furthermore, interpretable generative models would enable greater trust and confidence in sequences designed by such models, making them more viable for sensitive applications, such as drug design [63,64].

To elucidate the underlying determinants and biological principles that influence sequence generation by autoregressive generative models, we have developed a novel attribution assignment method (GAMA), which highlights the importance of specific positions for the positive class. GAMA is based on Integrated Gradients (IG) for LSTMs. We demonstrated GAMAs application to LSTMs trained on synthetic sequences and the results suggest that the importance of features gives us information on binding rules, i.e., amino acids that are important for binding. We define "Importance" as the difference between the integrated gradients (attribution assignment method) of a trained LSTM and a randomly initialized LSTM. We also show the applicability to antibody sequences.

Briefly, LSTMs and other state-space models estimate a probability distribution for the next token in a sequence given the previous tokens [65]. In this way, LSTMs model the entire sequence distribution globally and unsupervised while learning a supervised model locally for next-token prediction [66]. IG works on the local level of the autoregressive model, i.e., predicting a single element from the previous element, by identifying the impact of one input-output token pair [1]. By summarizing the IG results over all possible combinations of input-output pairs, GAMA obtains the global distribution that the model has learned.

Monoclonal antibodies (mAb) are used for various applications, such as drugs or as diagnostic biomarkers [63]. Each individual drug and application requires a specifically designed mAb. MAb design is currently performed either in vivo or in vitro [67]. This is usually very time and cost-intensive, for example, lead times to mAb discovery and design are on average >3 years [64,68]. Generative in-silico design of mAb sequences is a promising avenue for improving mAb design lead times and costs. Additionally, mAb sequences are a good modeling [44] system for evaluating our methods due to the availability of synthetic ground truth data [44]. The structural binding of an antibody to a target protein is described by a set of amino acids (termed paratopes) that interact with the antibody binding site of the target antigen (termed epitopes) [63]. Paratopes are typically localized in the complementarity-determining regions (CDR) of antibodies. Paratopes can have complex interactions with epitopes[63]. Consequently, attribution assignment methods must strongly consider the complexity [44] of the datasets and the robustness towards noise. Paratopes are usually gapped motifs [69] and we investigate the hypothesis that attribution methods will focus on binding residues preferentially, and, therefore, an analysis of attributions could be used to predict the paratope of antibody sequences. Successful attribution assignment of these motifs will pave the way for interpretable deep generative models [70].

In this work, we trained an LSTM with an autoregressive unsupervised objective on synthetic 1-class CDR3-antigen binding data at different levels of complexity as well as real-world data and identified the important features learned by the generative model using our newly developed method "Generative Attribution Metric Analysis" (GAMA).

# Results

## A method to quantify and highlight important motifs in generative models

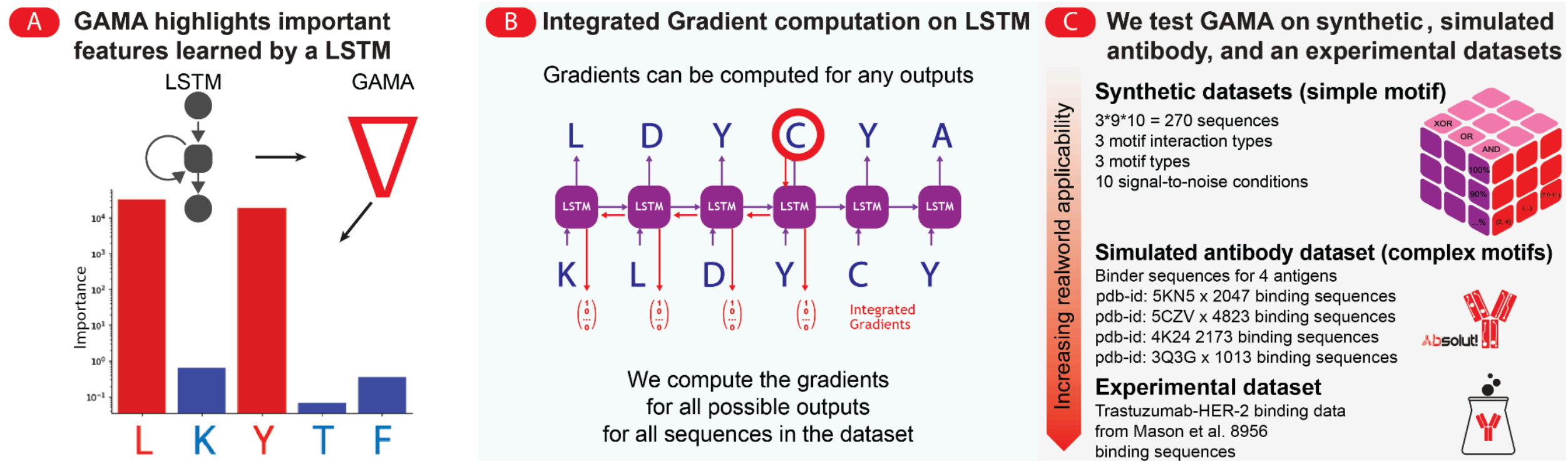

**Figure 2. Calculation of integrated gradients (IG) on LSTM models and experimental evaluation of GAMA.** (A) GAMA highlights important sequence positions that were learned by a generative LSTM model. We define "Importance" as the difference between the integrated gradients (attribution assignment method) of a trained LSTM and a randomly initialized LSTM. (B) The gradients can be computed for every output token at each sequence position w.r.t the input encoding. The figure illustrates the computation of gradients for the fourth "C" output; the red arrows indicate the computation flow of the gradients and the red tensors represent the gradient components of each input position. We compute the gradient components for several encodings that are obtained by linearly interpolating between the encoding of the input sequence and a "zero input baseline". The gradient components are pooled together to compute a global integrated gradient. They are computed for a trained LSTM with a sequence generated from the same LSTM over all outputs. GAMA consolidates all possible gradients into a statistic that highlights important sequence positions. (C) GAMA is evaluated on three distinct datasets with increasing levels of real-world applicability for antibody binding (270 synthetic datasets, Absolut!-simulated datasets, and Trastuzumab-HER2 binder dataset). We test our method by applying GAMA to models trained on datasets with known features.

Generative models have been used to learn and generate antibody sequences with desired properties. We develop a tool, GAMA, to extract information from trained generative models (Figure 2A). In particular, we identify the features learned by a generative model as follows: First, we apply interpretability/attribution approaches, such as IG, to the trained autoregressive generative model "main model" to obtain the importance of each input/output pair - e.g. the LSTM predicting the next amino acid based on the previous ones (Figure 2B). Subsequently, we repeat this process for a "reference model" and use the difference between the main and the reference model to obtain the learned features that are particular to the main model. See Algorithm 1 for details.

In the context where an antibody dataset contains binding sequences to an antigen, we hypothesize that if a generative model has learnt biological rules of antibodies, GAMA should reveal AAs important for the binding (Figure 2), and define the capacity to identify motifs as the predictive power of the model. We test our method by applying GAMA to models trained on datasets with known binding patterns. Since datasets with known binding patterns are rare, we first use 270 simple synthetic datasets with specific properties and implanted motifs to assess motif recovery under noise, motif position, inter-motif interactions, and motif length. We define motifs as short repeating elements of a sequence in a dataset. We assume this simulates important features of biological datasets that all share a single common gapped motif (paratope motifs are usually gapped motifs [69,71]). Then we evaluate GAMA on complex simulation-derived Ab-Ag binding simulations[44] with known binding motifs to assess if generative models can learn binding motifs. Lastly, we evaluate GAMA on an experimental dataset of HER-2 binding antibody sequences [72] (the Trastuzumab paratope is a non-gapped motif of size four) where we are only certain of the motif of one of the sequences (Figure 2C).

An LSTM is an autoregressive generative sequence model, which means it iteratively predicts the next element within the sequence given all previous sequences (next-character prediction), and is trained with teacher forcing. Such a model can sample novel sequences given a "start token".

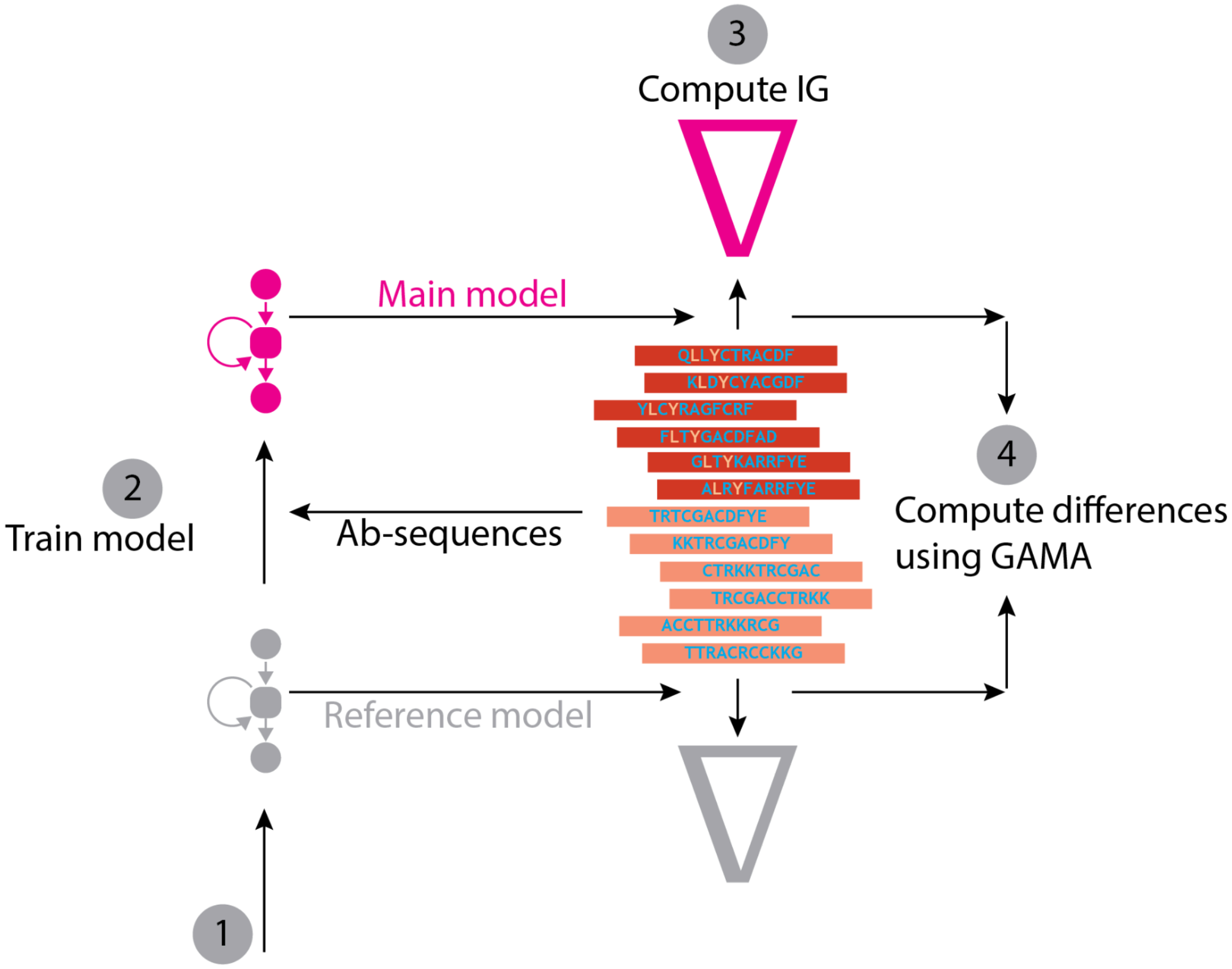

**Figure 3. GAMA assigns importance values to tokens within a sequence that quantify the influence a specific token has for the generation according to the generative LSTM.** Relevant signal motifs can be identified by comparing the IG of a trained model and the IG of a randomly initialized model. Here we use an LSTM to showcase the capability of our metric to assign attribution in autoregressive generative models. (1) An LSTM is initialized randomly and saved as the "reference model". (2) The reference model is trained generatively on antibody (Ab) sequences. The resulting model is then saved as "trained model". (3) For both models, we computed the IG-tensor (Figure 10) and reduced the dimension from 4D to 2D tensor. (4) We computed the similarity function for the two tensors (Equation 1).

The importance metric (GAMA) is derived by comparing IG signals from a randomly initialized reference LSTM and a trained LSTM model. Our similarity function M(t:=token position) is described in Equation 1. The similarity function measures the absolute median difference between the IG values for the reference model and the trained model. The difference was weighted by the normalized sum of the variance of the IG values in the reference model and the IG values in the trained model. In the next section, we will benchmark GAMA on a simple synthetic dataset, a more challenging antibody synthetic data [44], and antibody experimental data [72].

$$M(t) = \frac{|median(\mathcal{D}_{t,baseline}) - median(\mathcal{D}_{t,trainedModel})|}{\left(\frac{variance(\mathcal{D}_{t,baseline}) + variance(\mathcal{D}_{t,trainedModel})}{\sum_{i \in N} variance(\mathcal{D}_{i,baseline}) + variance(\mathcal{D}_{i,trainedModel})}\right)}$$

**Equation 1** describes the computation of the similarity function M(s) that computes the similarity of our trained model IG distribution and our reference model IG distribution at token position t. $\mathcal{D}_{t,trainedModel}$ indicates all IG datapoints computed on the trained model at sequence token position t. N indicates the total sequence length. A large difference indicates an important sequence position.

## Benchmarking of LSTM-based motif recovery across three key parameters: noise ratio, motif position, and motif complexity on synthetic datasets

We investigated how GAMA is affected by three variables: (i) noise ratio (ii) motif position (iii) motif logic. Noise ratio (i), is a ratio defined by the number of sequences containing motifs (motif_sequ) and the number of contaminating sequences not containing any motifs (no_motif): $\frac{motif_sequ}{motif_sequ + no_motif}$.
Motif position (ii) refers to whether the amino acids of a motif are concentrated towards the beginning, middle, or end of a sequence. Motif logic (iii) designates how the amino acids within a motif are allowed to interact to be counted as a motif. For example, a motif with “AND” logic must have all associated positions be set to the active, to be counted as a motif. We investigate three logic types: “AND”, “OR”, and exclusive or “XOR”.

To investigate the impact of these three parameters (parameter 3-tuples), we explored all possible combinations of the parameter 3-tuples. Each experiment consisted of a synthetic dataset as described in *Methods*. Briefly, the 3-tuples had the following values: three different logic motif interactions, three motif positions times three motif lengths, and ten signal-to-noise conditions, which resulted in 3*(3*3)*10=270 experiments (Figure 2C). An example of such a 3-tuple is a “L” implanted at position two, a “Y” implanted at position four, with a motif logic of “OR”, and a signal-to-noise ratio of 80% as seen in Figure 5. Since we implanted the artificial motif into these datasets, we know the ground truth motif position.

We investigated the usefulness of the GAMA importance values by applying it to LSTM models that were trained on datasets with known features and utilizing the importance values in a motif retrieval task. We demonstrated sufficient model convergence by sampling from the resulting models and plotting their frequency distribution (Supplementary figure 10). We quantify the performance of the GAMA importance value by comparing the sequence elements with the highest GAMA attribution with the known ground truth motifs. Retrieval performance is the ability of GAMA to recover relevant motif sequences accurately. Therefore, high retrieval performance indicates high interpretability. The noise ratio tests whether GAMA is robust to noise within the dataset; in biological datasets noise can be introduced, for example, through imperfect FACS sorting or library amplification or sequencing-related errors. The motif position parameter investigates if the retrieval performance is position invariant because it could be degraded due to gradient vanishing. The motif logic investigates the ability of our method to identify motifs with non-linear amino acid interactions.

## GAMA achieves high motif retrieval performance at high signal-to-noise ratios

We sought to understand how robust GAMA is in noisy datasets. Biological dataset creation is error-prone, i.e., sequences that are wrongly labeled as positive [73]. For example, FACS cell sorting can introduce noise when identifying high and low binding Ab binders, thus, introducing false positives into the dataset, we are not considering false negatives because we do not use negatively labeled data. Large numbers of sequences without motifs can also be inherent to certain datasets, for example, antibody repertoires sample all immune sequences from a subject with a disease status, but each subject will have many non-disease status-associated antigens.

We asked how the contamination of false positive sequences in a dataset impacted the capacity to identify motifs, by adding sequences with uniform randomly sampled amino acids to the dataset with different ratios (that we call the signal-to-noise ratio). The signal-to-noise ratio quantifies the number of noise sequences added to the 10 000 signal sequences, i.e., 100% signal-to-noise means only signal sequences and no noise sequences (10 000 signal sequences, 0 noise sequences) while a signal-to-noise ratio of 10% means 90% noise sequences (10 000 signal sequences, 90 000 noise sequences). Figure 5A shows an example of a dataset with signal-to-noise ratio of 80%.

We increase the signal-to-noise ratio from 0% to 100% in increments of 10 percentage points. For each signal-to-noise ratio we consider three separate motif-logic conditions (AND, OR, XOR) at each 9 different motif positions. This results in a total of 270 experiments (Figure 2 C). In Figure 4 we show the results for all experiments with signal positions two and four for all motif logic types without averaging over them. We can see that the GAMA attributions have the strongest distinction on the two and four positions for a signal-noise-ratio of 100% (no noise sequences) and “AND” motif logic. This distinction diminishes with more noise sequences and for “OR” and “XOR” logic datasets.

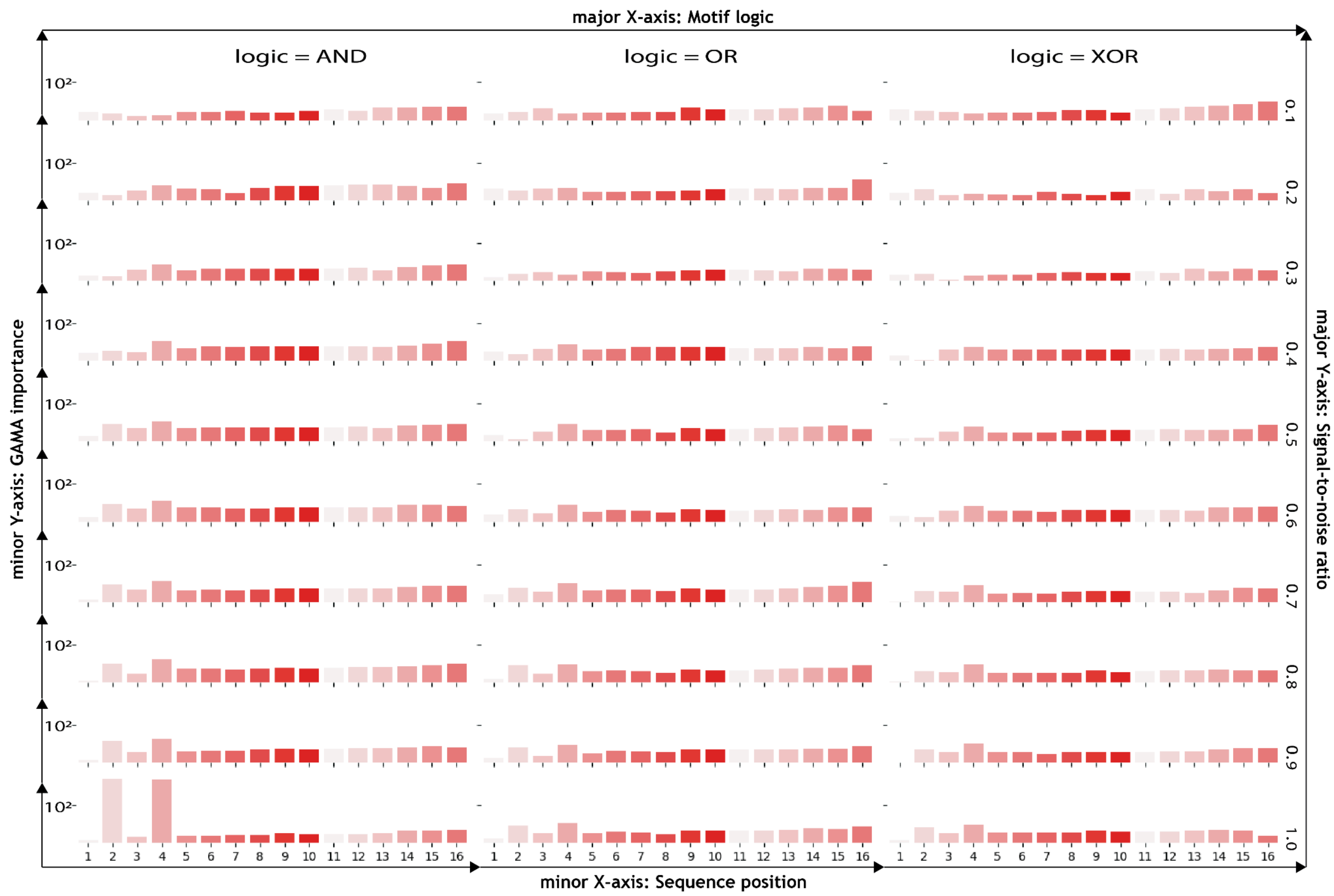

**Figure 4. GAMA attribution magnitude decreases with higher noise ratios.** Several basic examples are depicted in this figure where uniformly randomly sampled sequences were implanted with a motif for positions two and four with 10 000 sequences. We created 30 such datasets for all possible logic conditions and signal-to-noise-ratios. The plot compares the GAMA values on a log-scale (minor y-axes, left hand side) for all sequence positions (minor x-axes, at the bottom), between noise ratio (major y-axes, right hand side) and motif logic (major x-axes, on the top). We observe that the GAMA values at non-signal positions remain constant while the GAMA values at signal positions decrease with an increasing number of noise sequences.

False-negative rate quantifies the proportion of incorrectly retrieved instances, thereby providing insight into the effectiveness of GAMA in identifying true motifs. We averaged the false-negative-rates for 9 position experiments and plotted them with respect to each signal-to-noise-ratio for motifs that were formed by one of the logic conditions (Figure 5 B–D).

Since, to the best of our knowledge, there are no attribution assignment methods for 1-class sequence generative models available, we computed a random baseline with a Monte Carlo simulation to derive the false-negative-rate, one would get if the motifs were retrieved at uniform random; resulting in a false-negative-rate 0.93. GAMA outperformed the baseline in all experimental conditions. GAMA had the best performance for motifs with AND-logic, motifs were retrieved with 0% false-negative-rate for a signal-to-noise-ratio ≥80%. For a signal-to-noise ratio of 40%–70%, the signal is identified with a false-negative-rate lower than 50% (Figure 5 B). GAMA could still recover motifs with an OR-logic but the performance was worse for the entire signal-to-noise-ratio range. Motifs were retrieved with a false-negative-rate lower than 50% for signal-to-noise-ratio of 50%–100% (Figure 5 C). Motifs with Exclusive-OR logic are the most difficult to retrieve with GAMA. The false-negative-rate is worse for the entire signal-noise-ratio range and the false-negative-rate is below 50% only for signal-to-noise ratios of 90%-100%. To summarize, GAMA was able to identify the signal across all noisy dataset conditions to some extent. We investigated how effectively the GAMA function retrieves the motif positions, for all 270 datasets. Subsequently, we calculated and visualized the false-negative error rate (Supplementary figure 3), and the TopK (until full retrieval) (Supplementary figure 4).

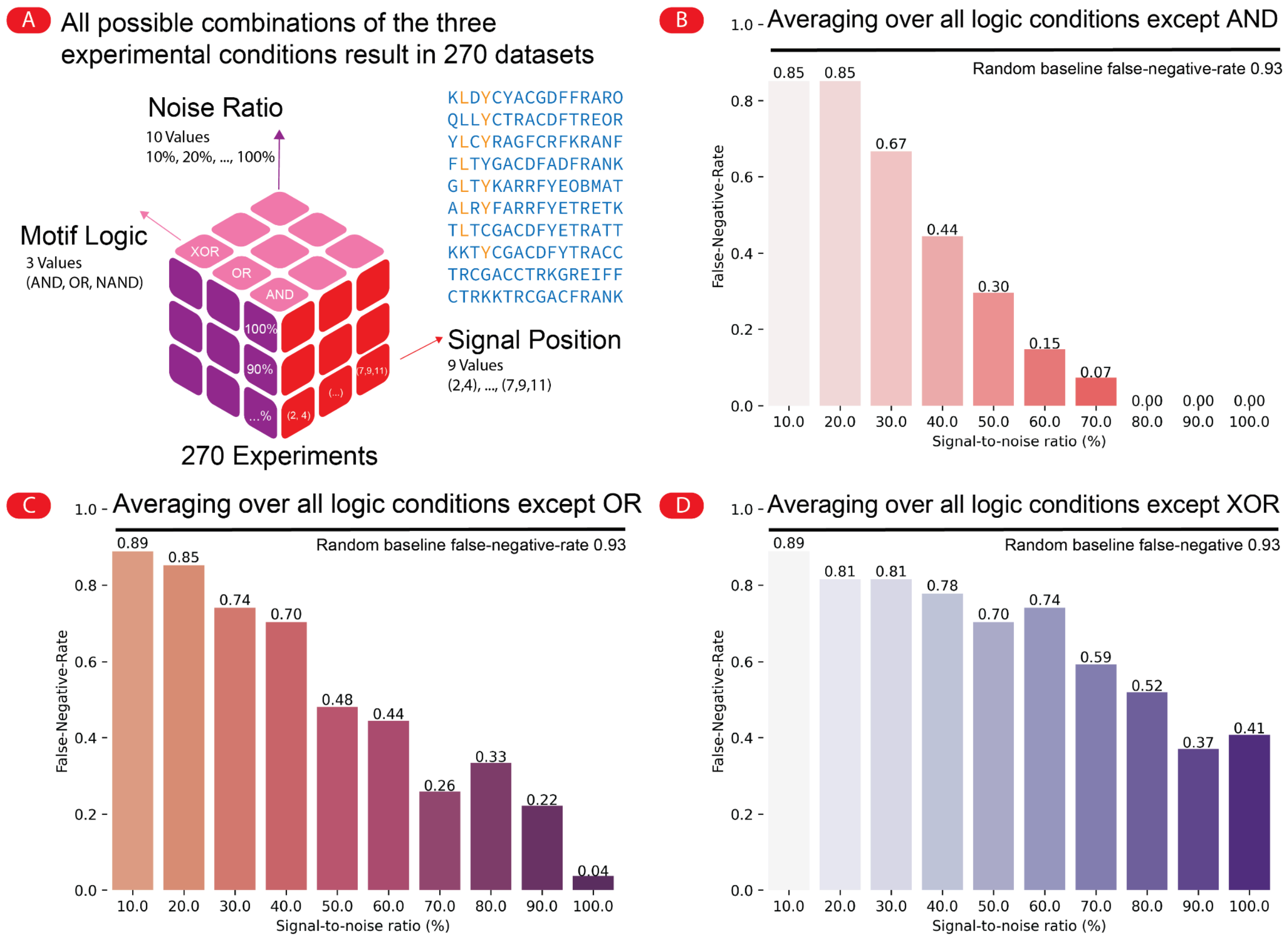

**Figure 5 GAMA false-negative-rate of motif retrieval is negatively correlated with the signal-to-noise ratio of the data.** (A) We defined 270 experimental conditions, with each experimental condition being a synthetically generated dataset, the sequences are sampled randomly first, then a motif is implanted into the non-noise sequences. Each dataset is generated by varying three key parameters: “Signal to Noise Ratio”, “Signal Position”, and “Motif Logic”, as visualized by the cube. An example dataset with OR logic, 80% signal-to-noise ratio, and (2, 4) signal position is visualized. For each possible combination of parameters, we create one dataset consisting of 10 000 sequences containing the motif and an appropriate amount of noise sequences without the motif, depending on the “Signal to Noise Ratio”-condition. Each parameter has an associated false-negative-rate. The false-negative-rate is defined as the ratio of motif elements that were correctly identified by the k-highest GAMA positions. A random baseline is computed that indicates the average false-negative-rate if the top-k motifs positions are identified randomly. (B) The false-negative-rate decreases with higher signal to noise ratio for motifs that are connected by AND logic. For signal ratios of ≥80%, we obtained perfect results (0 false-negative-rate). (C) The false-negative-rate decreases with higher signal to noise ratio for motifs that are connected by OR logic. OR-logic motifs perform worse than AND-logic motifs but better than Exclusive-OR-logic. (D) The false-negative-rate is above 50% for all signal-to-noise conditions ≥80%. The false-negative-rate decreases with higher signal-to-noise ratio for motifs that are connected by Exclusive-OR-logic (XOR). Motifs connected by XOR-logic are the most challenging to recover. The numbers on the bar charts signify the false-negative-rate.

## The performance of GAMA is position-invariant

We investigated whether the performance of our method is position-invariant, as motifs may vary in positions across biological sequences. Since our method was designed for autoregressive models, motif positions at the beginning of a sequence may be less reliably identified than those toward the end. The weaker gradients experienced for elements located towards the end of the computation graph likely contribute to this discrepancy [74]. To examine the reliability of motif identification, we implanted motifs at three different positions within the sequence: the front, middle, and end. We averaged the false-negative rate over the three motif positions (Figure 6, Supplementary figure 5). We observed very similar false-negative-rate distributions for all three motif positions. Only slight shifts in the distributions are observable. Therefore, we concluded that the performance of GAMA is position-invariant.

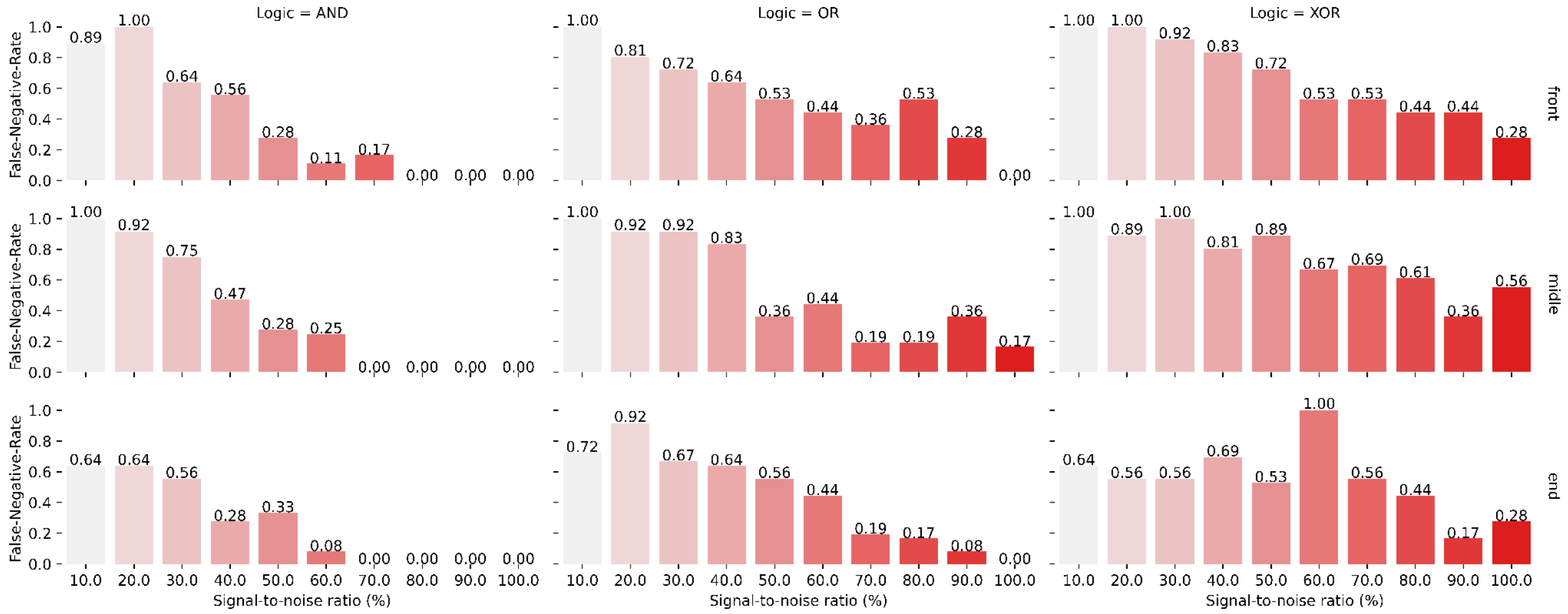

**Figure 6 Retrieval performance of motifs is position invariant.** The numbers on the bar charts signify the false-negative-rate on the minor y-axis, signal-to-noise ratio on the minor x-axis. Major x-axis shows the results for the three motif-logic variants (**AND**, **OR**, and **XOR**). The major y-axis compares motifs at the **front**, **middle**, and **end**. They are obtained by averaging over different motif lengths at the front, middle, and end. We can see that motif recovery rates are similar for the three position conditions.

## The length of the motif correlates positively with the GAMA false-negative-rate

We investigated whether longer motifs render it more challenging to recover the motif accurately. For instance, Akbar and colleagues have shown that antibody binding sites (paratopes) can vary in amino length [63,71]. Multiple relevant elements in our sequence may dilute the IG signal. To answer this question, we queried motifs of different lengths (binary, trinary, quaternary). Binary motifs have a lower false-negative rate than the larger motifs and the false-negative-rate is similar across all three motif lengths. (Supplementary figure 6). This can be explained by the fact that larger motifs also have more possible combinations.

## GAMA false-negative-rate of retrieved motifs increases in the following order: AND-logic, OR-logic, XOR

The function of a biological sequence is driven by inter-position dependencies. Individual amino acids within binding motifs of CDR sequences can have different effects on the binding affinity. For example, a CDR might only effectively bind an antigen if all amino acids of the motif have the correct binding structure (AND-interaction). In contrast to a binding motif that just requires one of the amino acids to be correct (OR-interaction). Different interactions between amino acids within the motif can be described by logical operations. We investigated the performance of our method on different motif interactions. We considered three logic operations AND, OR, and XOR. Valid signal sequences must conform to this logic otherwise they are considered noise sequences. For example, for the AND logic to be considered valid, all signal positions must be activated, but for the OR logic only one of the sequence positions must be activated. We also included longer signals that contain more than two sequence positions; we include binary, trinary, and quaternary. The non-binary sequences are created by chaining the respective logic operation (i.e., s1 AND s2 AND s3). The false-negative-rate rate for all experiments with the same signal logic is averaged. The results are plotted in (Figure 5; A "AND-logic", B "OR-logic", C "XOR-logic"). In summary, we evaluated the performance of our method across various motif interaction logics (AND, OR, XOR) and extended analyses to include longer signals, demonstrating how different logical dependencies influence false-negative rates.

## GAMA attributions on synthetic antibody binding data correlate with positional contribution to binding affinities

In the preceding chapters, we delineated the performance characteristics using a simplified synthetic data set (Figure 5). Following this, we assessed the performance of GAMA on a more complex and realistic synthetic dataset, which we created using antibody binding simulations generated with the Absolut! Framework [44]. This framework provides binding affinity of antibody CDRH3 sequences together with the amino acid-wise binding energy values contributing to the total binding affinity [45], serving as a proxy for the position-specific ground truth amino acid significance, which is largely unattainable in experimental data. The informativeness of binding locations is likely correlated with the binding energy if a specific binding motif mediates the physical interaction with the epitope. It has to be noted that amino acids not within the paratope may also have high information content, for example, an amino acid that causes a change in the 3D structure of the paratope.

The Absolut! framework discretizes an antigen and systematically explores all possible CDRH3 binding configurations and computes the best binding affinity (Figure 7A). We investigated whether GAMA effectively highlights amino acid motifs that correspond to the binding energy contributions of each amino acid in the CDRH3 region.

We generated a set of binding CDRH3 sequences for four different antigens (pdb-id 5CZV: 4823 CDRH3 sequences (Figure 7C), pdb-id 5KN5: 2047 CDRH3 sequences (Supplementary figure 7), pdb-id 4K24: 2173 CDRH3 sequences (Supplementary figure 8), pdb-id 3Q3G: 1013 CDRH3 sequences (Supplementary figure 9)). We selected these CDRH3 sequences from the top 1% binders to a specific epitope on a given antigen, containing at least 1000 sequences with the lowest entropy in the dataset. (Information)-entropy quantifies the uncertainty within a dataset; the lowest entropy dataset is the one with the highest information.

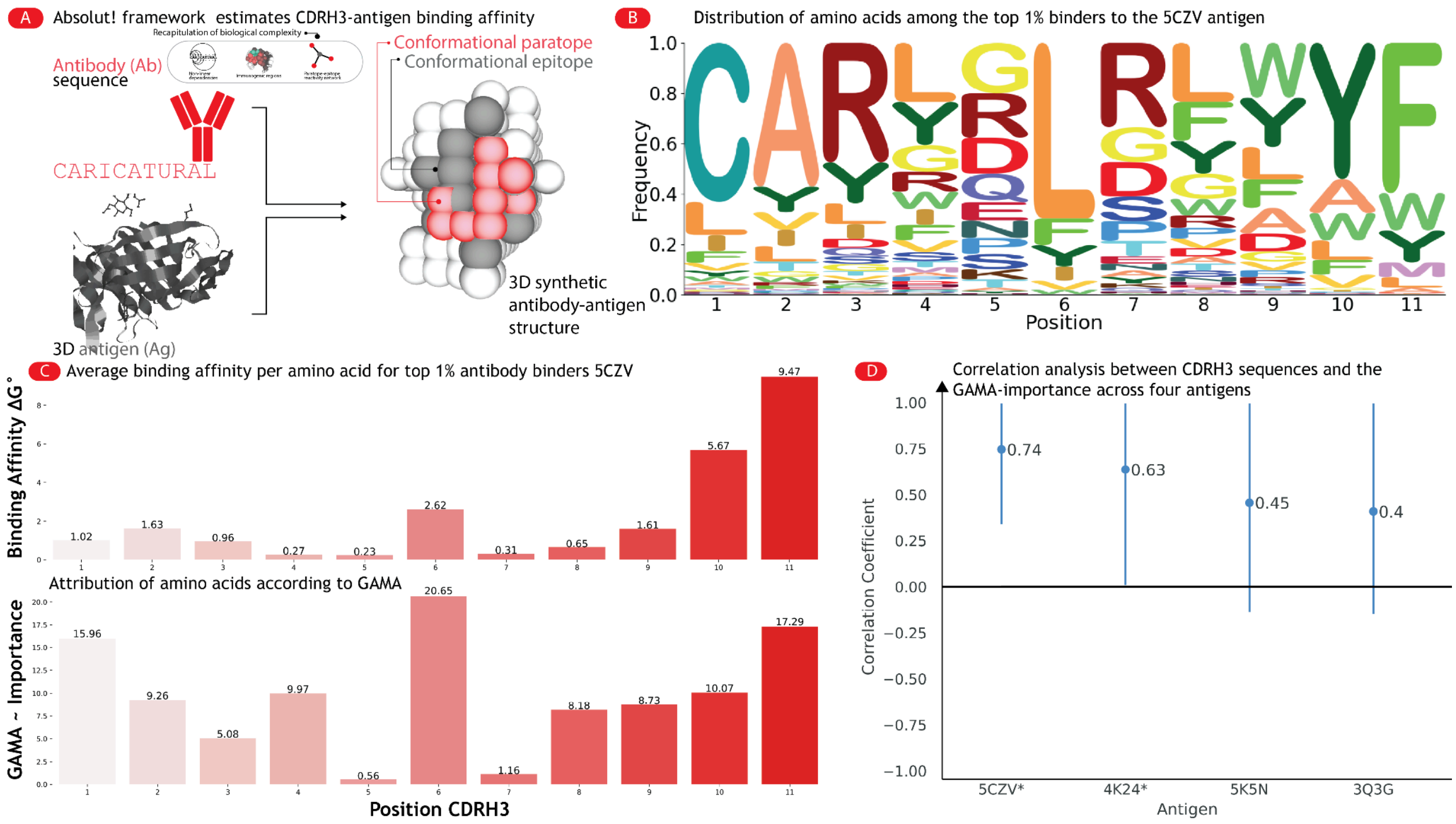

**Figure 7 Synthetic amino acid binding affinities correlate with GAMA assigned attributions** (A) The Absolut! framework estimates CDRH3-antigen binding affinity between CDRH3 sequences and antigens by discretizing the antigen's structural conformation and exhaustively exploring all potential binding configurations for the CDR sequence. (B) Distribution of amino acids among the top 1% binders to the 5CZV antigen. (C) The upper bar chart illustrates the average amino acid binding affinities of 4823 CDRH3 sequences, representing the top 1% binders to the 5CZV antigen as per Absolut!. The lower bar chart shows the corresponding attributions assigned by GAMA, with a Spearman correlation between the CDRH3 sequences and the GAMA-importance, of 0.74. This correlation is significantly greater than zero, with a one-sided p-value of 0.004 (<0.05). (D) Spearman correlation analysis across four antibody datasets, each created with a different antigen, compared against the Absolut! ground truth. 95%-confidence intervals, computed by bootstrap sampling (1000 samples), are displayed as lines. Significant correlations are indicated with an asterisk (*), the significance cut-off is **α**=0.05.

For the 5CZV derived dataset (frequency distribution plotted in Figure 7B), we compared the Abolut! [44] ground-truth data with the attribution assignments derived from GAMA. Our analysis successfully identified a high binding affinity amino acid at position six, as shown in Figure 7C. To evaluate the consistency across datasets, we calculated the Spearman correlation between Absolut! binding affinities and GAMA attributions for all four datasets, marking statistical significance with asterisks (*) in Figure 7D. The non-perfect correlation may be explained by the fact that most informative locations should not necessarily be binding locations, for example, an amino acid that causes a change in the 3D structure of the paratope. Additionally, we computed one-sided 95% confidence intervals for the correlation coefficients using bootstrap sampling, also depicted in Figure 7D. The one-sided p-values for the four antigens were as follows: 5CZV p-value = 0.004 < 0.05; 4K24 p-value = 0.018 < 0.05; 5K5N p-value = 0.08; 3Q3G p-value = 0.196. To summarize, our analyses highlight significant correlations between Absolut! binding affinities and GAMA attributions for two antigens (5CZV and 4K24), with statistical support provided by one-sided p-values and confidence intervals, while variability in other datasets underscores the complex relationship between binding locations and structural influences.

## GAMA identifies the most important amino acid positions in the CDR3H region of the Trastuzumab antibody

We demonstrated that GAMA effectively identifies significant motifs in synthetic sequences and synthetic antibody sequences, even under noisy conditions. Notably, GAMA achieved best performance with datasets exhibiting less than 50% noise ratio and motifs connected by "AND"-logic. Although most other conditions surpassed the baseline, motif identifications remained prone to errors (Figure 5). To extend our analysis, we examined the experimental Trastuzumab-HER-2 binding data from Mason et al. [72] (hereafter referred to as the "Mason dataset"). The Mason dataset comprises of 8955 sequences that were generated by mutating the original trastuzumab CDRH3 and filtering them for sequences that bind the HER-2 antigen; the resulting frequency distribution is visualized in Figure 8A.

Given the experimental nature of the dataset, the amount of noise in it is unknown. That said, the crystal structure of unmutated Trastuzumab bound to HER-2 (pdb-ID: 18NZ) identifies the key paratope amino acids for binding at positions 102, 103, 104, and 105 (Figure 8B). It remains uncertain whether these positions maintain their significance within the Trastuzumab variants. Since GAMA identifies averages over all sequences, different binding motifs within the same dataset may dilute the GAMA attributions across the sequence length. Therefore, the correlation between the informativeness and the binding motif might not be perfect.

Figure 8 presents our findings with a logo plot of the amino acid frequency distribution shown in Figure 8A. The GAMA activations for the Mason dataset, highlighted in Figure 8C, underscore positions 103, 104, 105, and 107 as particularly important for the generative model, with some overlap between the predicted and known binding positions of the original Trastuzumab-HER-2 interaction. In summary, the analysis in Figure 8 reveals key positions identified by GAMA activations that align partially with known binding sites, emphasizing their relevance to the generative model and the Trastuzumab-HER-2 interaction.

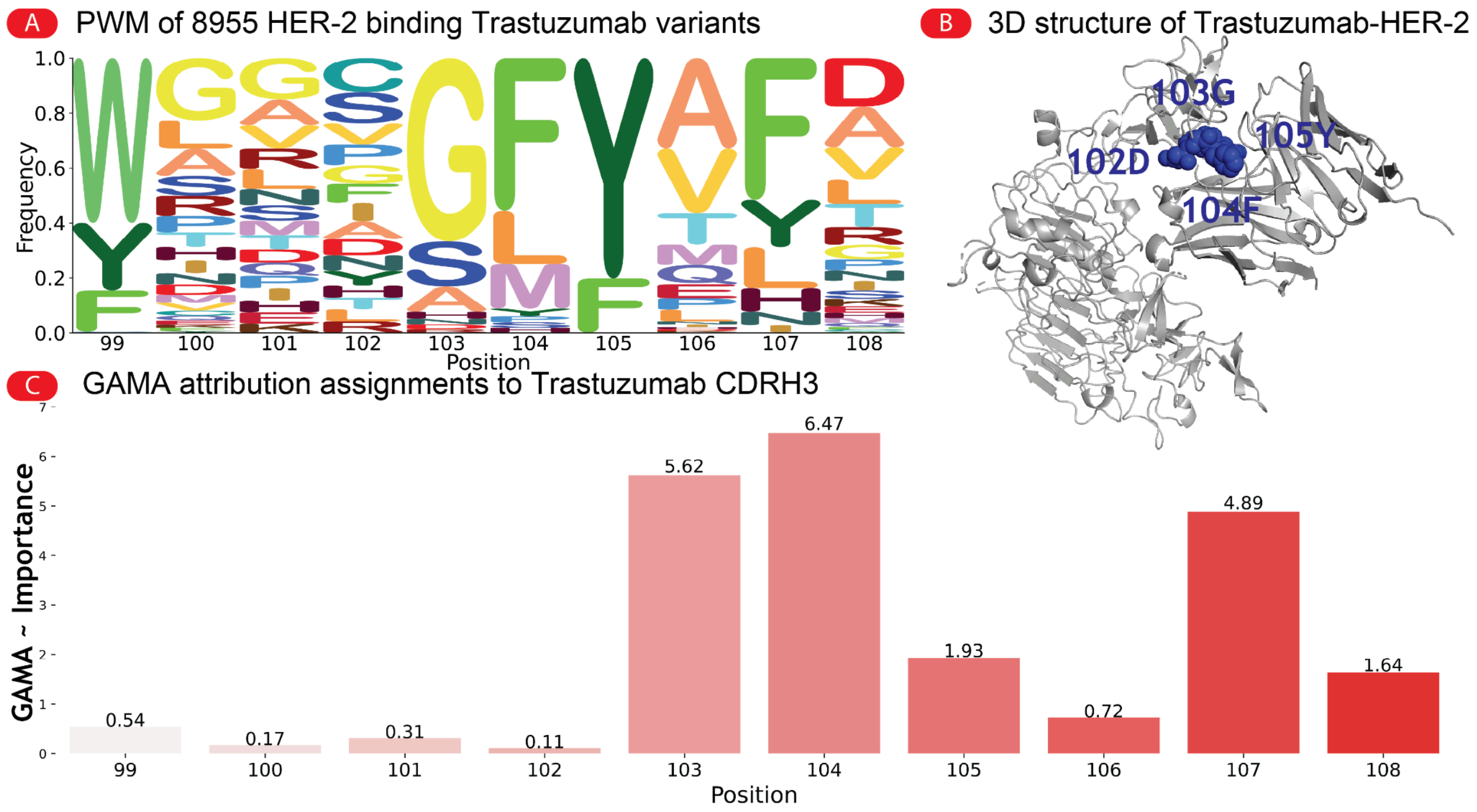

**Figure 8: GAMA attributions identify three out of four important amino acids positions of the unmutated HER-2/Trastuzumab binding data** (A) Logoplot of the 8955 sequences of the mason dataset. (B) Trastuzumab-HER-2 binding structure (pdb-ID: 1N8Z), highlighting the four binding amino acids of the original trastuzumab sequence. (C) Position (x-axis) attributions according to GAMA-importance (y-axis).

# Discussion

In this paper, we developed a novel analysis method, GAMA, to enhance the interpretation of generated sequences for autoregressive generative models and ameliorate the effects of negative label bias. This advancement is particularly relevant for datasets devoid of negative label sets, known as "Positive and Unlabeled data" (PU-data), which find extensive applications beyond biological contexts, including recommender systems, fraud detection, information retrieval, and personal advertisement. Methods for PU-data typically employ semi-supervised strategies within an iterative "Two-step" learning process, wherein a supervised method classifies unlabeled data, subsequently using the new labels as training data for successive iterations [75–77]. While these PU-data approaches are versatile, they do not provide solutions specifically tailored for generative models.

Recent generative approaches to antibody sequence design often fail to elucidate why a generative model considers a newly generated sequence to be optimal for a specific task [78,79]. GAMA addresses this gap by offering interpretability, and it is applicable to a wide range of (differentiable) autoregressive models, including large language models (e.g. protein language models), which have demonstrated the most promising recent advances in generative antibody design [80,81], as well as other enhancements in state space models such as xLSTM [82], and Mamba[2]. GAMA could also be extended to address non-biological challenges, such as generative models for outlier detection employed in fraud detection, where the interpretability of model outputs is of importance.

Utilizing simple synthetic sequences with well-defined motif signals, we observed that GAMA can recover binding motifs under various sequence noise conditions and remains invariant to motif position (Figure 5, Figure 6). Our results indicate a strong correlation between GAMA and the ground-truth data in simulated antibody binding data (Figure 7D). We also demonstrated the efficacy of our approach using an experimental antibody binding dataset, thus confirming its applicability to real-world data (Figure 8).

The evaluation of GAMA on experimental HER-2/Trastuzumab binding data revealed that positions 103, 104, 105, and 107 are particularly significant for the generative model. When comparing these results to the four binding positions in the unmutated Trastuzumab sequence, which indicate positions 102, 103, 104, and 105 as most important for binding with the HER-2 antigen (Figure 8), we matched the positions 103, 104, 105 and 107, but failed to match position 102. This discrepancy between structural information and GAMA-identified positions could stem from several factors. Firstly, structural information is available only for the unmutated Trastuzumab sequence, making it uncertain to what extent mutated sequences retain the same binding amino acids. Secondly, GAMA provides average amino acid attributions over all sequences, meaning that multiple different motifs within the dataset could aggregate into a single "combination" motif. Thirdly, as demonstrated in Figure 5 with synthetic sequences, error increases with higher noise levels and more complex inter-position interactions.

In this study, we benchmarked GAMA using simulated paratope motifs of up to four interacting amino acids, separated by gaps of up to one or two amino acids. These motifs encompass the most probable types in terms of length and number of gaps. The distributions of paratope motifs have been thoroughly investigated [69]. For instance, the CDRH3 region exhibits a broad range of motif lengths, from one to fifteen, with a median of five, while gap numbers range from zero to five, with a median of one. In contrast, the CDRL2 region features paratope motifs ranging from one to seven in length, with a median of two, and gap numbers ranging from zero to five, with a median of two.

Applying GAMA to generative antibody sequence models represents a promising avenue for rational antibody design [25,45,83]. Understanding the rules governing antibody-antigen binding, which operate on the antibody and (protein)-antigen sequences [84], could facilitate the creation of novel, fit-for-purpose antibodies. Our work represents an initial step towards this goal.

Currently, GAMA analyzes single positions within the sequence chain, which hampers its ability to identify more complex binding patterns, such as higher-order dependency patterns [85], and may result in the erroneous identification of important positions [69]. GAMA offers an aggregated summary of the generated sequence patterns by integrating attributions across the entire dataset. This methodological approach, however, limits GAMA's capacity to discern motifs in environments with very low signal-to-noise ratios, thereby constraining its applicability to domains where motifs are anticipated to be sparse, such as Adaptive Immune Receptor Repertoire (AIRR) datasets. For example, only approximately 10% of TCR sequences in cytomegalovirus (CMV)-infected patients are associated with CMV [86].

Future work will focus on extending GAMA to single sequence analysis, enabling attributions to be computed for individual sequences. Currently, the noise of the integrated gradients for a single sequence is too high to allow for meaningful individual interpretation, necessitating the averaging of attributions over many sequences, but the gradients contain much more information in 4D, which should be further analyzed. Due to computational limitations, we only demonstrated the applicability of GAMA for LSTMs. Future research should demonstrate the applicability to modern sequence models such as transformers, and state-space models. Additionally, future research should incorporate pairwise and higher-order dependency information, as we know that CDR binding is mediated by the interaction of multiple amino acid positions within the CDR [44,72].

# Methods

## Datasets

| Dataset | Number of sequences | Reference |
|---|---|---|
| Synthetic dataset x270 | 10 000 per dataset | This manuscript |
| Synthetic antibody dataset (Absolut!) | Binder sequences for 4 antigens (pdb-id: 5KN5 x 2047 binding sequences, pdb-id: 5CZV x 4823 binding sequences, pdb-id: 4K24 2173 binding sequences, pdb-id: 3Q3G x 1013 binding sequences) | Robert et al. 2021 [44] |
| Experimental dataset | Trastuzumab-HER-2 binding data from Mason et al. 8956 binding sequences | Mason et al. 2021 [72] |

Table 1**. Datasets utilized for evaluation of GAMA.** We evaluated GAMA on three types of datasets. (1) 270 synthetic sequences that were systematically created to evaluate GAMA for robustness to noise, motif position invariance, and motif logic. (2) Data from the antibody simulation framework Absolut! [44] to demonstrate the viability of GAMA under complex conditions. For each antigen, we selected the top 1% best binding CDRH3 sequences based on energy and filtered them for sequences with one paratope-binding interaction. (3) Experimental mAb HER-2 binder sequences from the Mason dataset [72].

### Synthetic dataset generation

Whether CDR sequences are associated with binding to an antigen usually depends on short sequences of amino acids that physically interact with the epitope. To simulate this interaction, we created datasets containing sequences with uniform sampled amino acids, and containing implanted signal amino acids that designate a motif. Each dataset is associated with an experimental condition that is described by a 3-tuple: motif positions (list of motifs within the dataset, represented as a list of all sequence positions where the motif is implanted: (2, 4), (7, 9), (13, 15), (2, 4, 6), (7, 9, 11), (12, 14, 16), (2, 3, 4, 5), (6, 7, 8, 9), (11, 12, 13, 14)), motif logic (“AND”, “OR”, “XOR”), and noise ratio (1, 0.9, 0.8, 0.7, 0.6, 0.5, 0.4, 0.3, 0.2, 0.1). An example of how the dataset creation is structured can be seen in Table 2. Noise ratio is the ratio between sequences with an implanted motif and noise sequences devoid of any motif if motifs are present by chance the sequence is rejected and resamped. The randomly generated sequences without signals were inserted into the datasets to assess the model's robustness against noise. Specifically, the noise sequences were composed of uniformly sampled tokens, while the signal sequences were also uniformly sampled but included implanted signal tokens representing the motif. A token represents an element of the sequence and corresponds to an amino acid.

The synthetic datasets were generated using a deterministic seed to ensure reproducibility. Each sequence had a length of 16, a value chosen because it approximately represents the median length of human CDRH3 sequences [87]. Each experiment consisted of 10 000 signal sequences paired with a specific quantity of noise sequences. We demonstrated that 10 000 sequences are adequate for achieving robust model performance by training deep generative models on iteratively smaller subsets of a complete dataset (Supplementary figure 11).

We generated new datasets based on three primary variables: signal-to-noise ratio, signal positions, and signal logic. The signal position parameter also included motifs of varying sizes to examine the signal recovery rate of our model in relation to motif size.

| | | | | | | | | |
|---|---|---|---|---|---|---|---|---|
| (2, 4)<br>AND<br>1.0 | (7, 9)<br>AND<br>1.0 | (13, 15)<br>AND<br>1.0 | (2, 4, 6)<br>AND<br>1.0 | (7, 9, 11)<br>AND<br>1.0 | (12, 14, 16)<br>AND<br>1.0 | (2, 3, 4, 5)<br>AND<br>1.0 | (6, 7, 8, 9)<br>AND<br>1.0 | (11, 12, 13, 14)<br>AND<br>1.0 |
| (2, 4)<br>OR<br>1.0 | (7, 9)<br>OR<br>1.0 | (13, 15)<br>OR<br>1.0 | (2, 4, 6)<br>OR<br>1.0 | (7, 9, 11)<br>OR<br>1.0 | (12, 14, 16)<br>OR<br>1.0 | (2, 3, 4, 5)<br>OR<br>1.0 | (6, 7, 8, 9)<br>OR<br>1.0 | (11, 12, 13, 14)<br>OR<br>1.0 |
| (2, 4)<br>XOR<br>1.0 | (7, 9)<br>XOR<br>1.0 | (13, 15)<br>XOR<br>1.0 | (2, 4, 6)<br>XOR<br>1.0 | (7, 9, 11)<br>XOR<br>1.0 | (12, 14, 16)<br>XOR<br>1.0 | (2, 3, 4, 5)<br>XOR<br>1.0 | (6, 7, 8, 9)<br>XOR<br>1.0 | (11, 12, 13, 14)<br>XOR<br>1.0 |

Table 2 **Demonstration of synthetic dataset generation.** The table has 27 cells, each specifying a 3-tuple corresponding to a dataset for one of the experimental conditions. The table describes all datasets with a signal to noise ratio of 1.0. There are 10 tables like this for each signal-to-noise ratio resulting in 270 datasets in total.

## Synthetic antibody dataset generated with the Absolut!-framework

We evaluated GAMA using datasets obtained from the Absolut! framework. We generated four datasets where each dataset consists of CDRH3 sequences that bind a specific antigen: 5KN5 with 2,047 binding sequences, 5CZV with 4,823 binding sequences, 4K24 with 2,173 binding sequences, and 3Q3G with 1,013 binding sequences. Each dataset was generated by first identifying an epitope hotspot on an antigen, followed by selecting the top 1% of antibody binders to form the dataset. We conducted a comprehensive search of all possible epitopes within the Absolut! database and selected the four epitopes that yielded datasets with the lowest entropy (i.e., highest information content), while ensuring that each dataset contained a minimum of 1,000 sequences.

## Experimental dataset

The original dataset from Mason et al. [72], comprising HER-2 binders. The positive binder sequences were filtered to include only sequences with a minimum of two reads. This filtering step was undertaken to eliminate false negatives that may have arisen due to erroneous reads or contaminations (Figure 9).

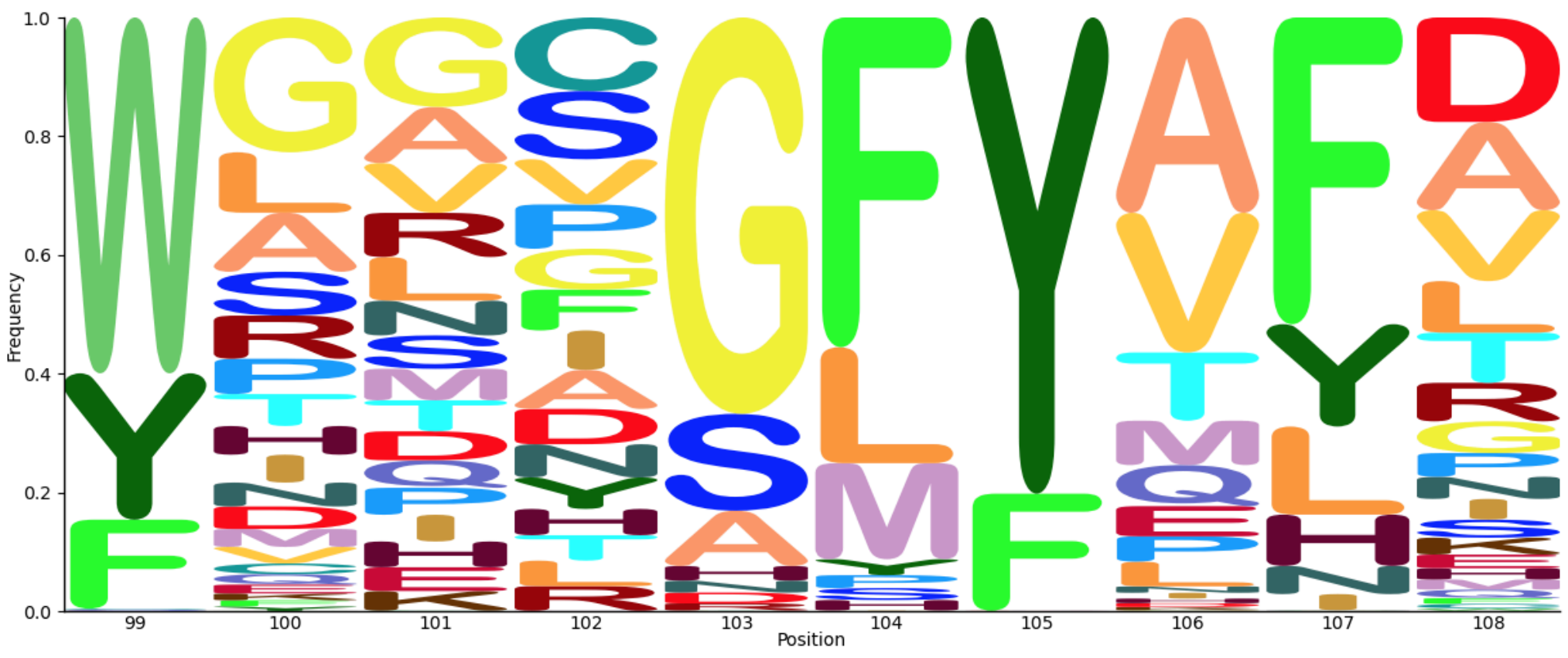

**Figure 9 Frequency distribution of amino acids across the position of the CDR sequences from Mason et al.**

## Computational methods and data processing

### LSTM training and sampling

We trained an LSTM (Long Short-Term Memory) [1] model to learn a dataset in a generative manner, enabling the sampling of novel sequences. Each sequence in a dataset was encoded using one-hot encoding while adding start and stop tokens. All sequences shared the same length within one dataset. The LSTM model is trained to predict the identity function, which allows the model to predict the subsequent amino acid in the sequence, a technique known as teacher forcing [66]. This allows the sampling of novel sequences from the same distribution. We sampled new sequences from the model by initiating the trained model with a start token, then progressively sampling one amino acid at a time until an end token was reached.

The LSTM model has a hidden state size of 1024. Training was conducted using teacher forcing and mini-batches, with the ADAM optimizer [88] from PyTorch 1.8.1, utilizing default parameters except for a learning rate set to lr=0.00001. This learning rate was determined through multiple test runs to ensure stable convergence (see Supplementary figure 1 and Supplementary figure 2). The weights were initialized with a uniform distribution according to $\mathcal{U}(-\sqrt{k}, \sqrt{k})$ where $k = \frac{1}{size_hidden_state}$. To ensure reproducibility of our stochastic experiments, the random seed value was set to zero. Figure 10 illustrates the prediction step of an LSTM, with input values shown in blue and output values in black, and gradients in black.

### Integrated Gradient analysis

Integrated Gradients (IG) [35] is a method for attributing the contribution of input features to the output of a machine learning model. Given an input, a model, and its respective output, IG quantifies the importance of each feature by measuring the degree of change in the output when that particular input feature is altered, as defined by the gradient. The importance is defined by the magnitude of change in the output resulting from a change in the feature.

Applying IG to generative models is generally non-trivial, as most attribution methods focus on explaining changes in predictions with respect to inputs, while deep generative models often lack a supervised objective that directly corresponds to the sample output. However, in the case of LSTMs, which are trained in a supervised manner to learn the identity function, IG can be computed with respect to the inputs for a single output.

IG were computed on the input sequence $x$ (encoded as a tensor with dimensions 22 x sequence length) and a zero input baseline $x\prime$. The zero input baseline mirrors the dimensions of the one-hot encoded input sequence but sets all values to zero, despite this not constituting a valid one-hot encoding and does not refer to a specific token. We generated 1000 equally spaced points along a linear path between the input sequence and the input baseline, with linear interpolations performed in the one-hot encoding space.

Each input interpolation is described as $x\prime + \alpha * (x - x\prime)$ for 1000 equally spaced $\alpha \in [0, 1]$. For each of the 1000 input interpolations, we used the LSTM model to predict outputs and subsequently computed the gradients with respect to the corresponding inputs. By averaging over all 1000 gradients, we obtained the IG for a given sequence. These gradients were computed using automatic differentiation provided by PyTorch 1.8.1.

### Calculation of integrated gradients and structuring into a 4D Tensor

The previous section described how IGs are computed. For this study, the IG are computed for each output token of a sequence, structuring the results into a four-dimensional tensor (1. Input encoding, 2. Input sequence token, 3. Output probability vector, 4. Output sequence token), which we hereafter refer to as the "IG-tensor" (refer to Figure 10). Given that IG can only be computed for a single output of the LSTM at a time, the LSTM, trained on a positively labeled dataset, was queried for the IG of an input sequence from the same dataset. This computation results in a two-dimensional tensor for a single output, representing the IG with respect to all LSTM inputs. The dimensions of this tensor are 22 (20 amino acids + start token + stop token) by the sequence length.

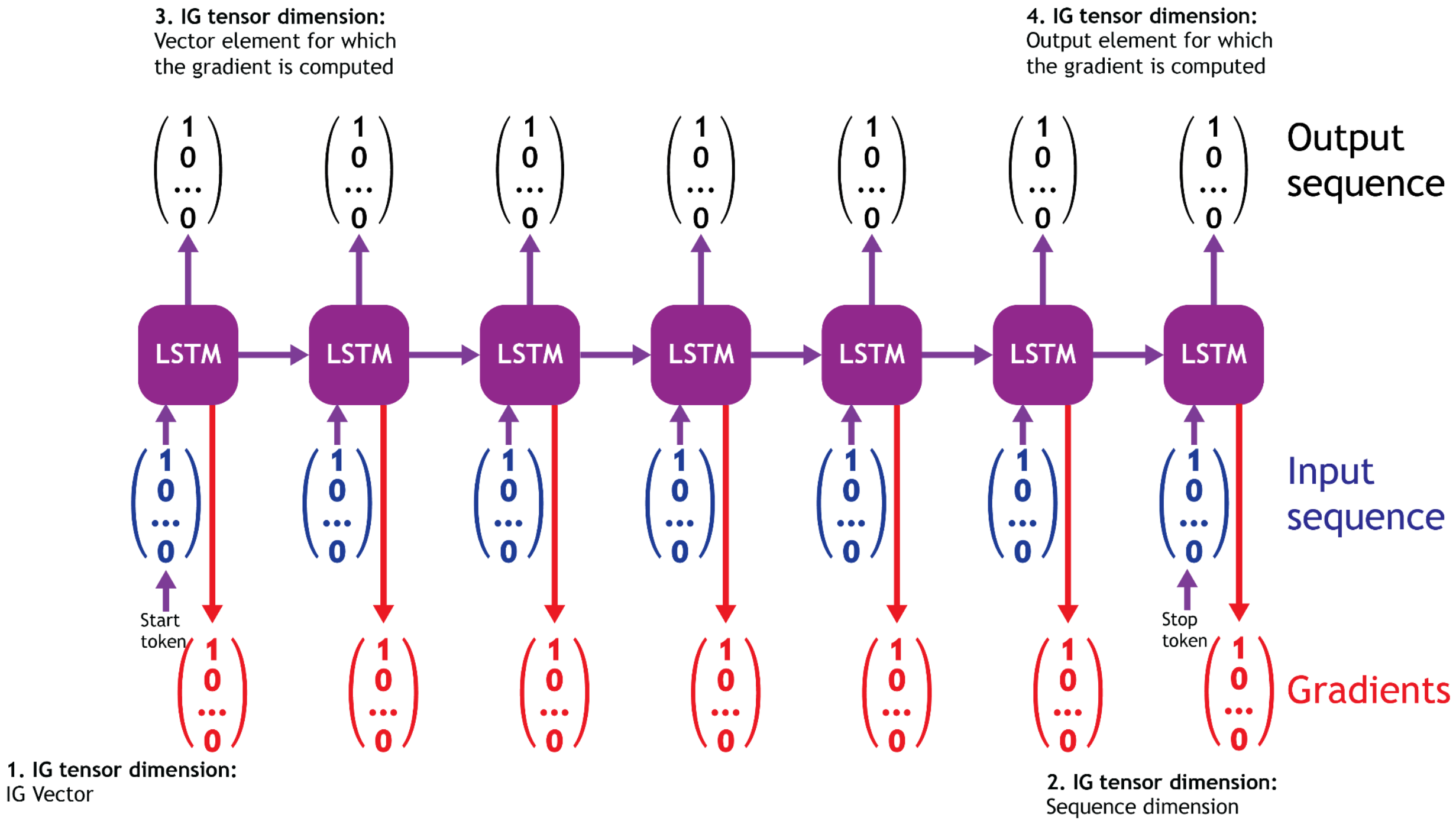

**Figure 10.Calculation of integrated gradients (IG) tensor:** The LSTM model is queried to compute the gradient (red)IG for a given input sequence encoded with one-hot-encoding (blue)., The IG is computed from the gradient, resulting in a four-dimensional IG tensor for each sequence. The dimensions of this tensor are as follows: **(1)** Integrated gradients are computed for all 22 elements (comprising 20 amino acids, a start token, and a stop token) within the encoding vector for each amino acid in the sequence. **(2)** A sequence vector corresponding to the length of the input sequence. **(3)** Gradients are calculated for each dimension in the output encoding of the softmax, with the output sequence length matching the input sequence length. **(4)** Gradients are computed for all elements in the output sequence.

We performed this IG computation for all possible outputs. Each output comprises the softmax output encoding of size 21 (20 amino acids + stop token) and a prediction for each token in the input sequence of size matching the sequence length. Consequently, this results in a four-dimensional IG-tensor with dimensions [22 (number of input tokens), sequence length, 21 (number of predicted outputs), sequence length]. Following the enumeration of all possible IG computations, we reduce and summarize the 4D IG-tensor to a more manageable two-dimensional tensor in the subsequent section.

First, the 4D-IG-tensor was reduced to a 2D tensor by eliminating the first dimension (input encoding) and the third dimension (the output dimension for which the gradient is computed). The first dimension of the IG tensor was reduced by only selecting scalar values that correspond to the encoding position of the input amino acid as other positions are zero. The third dimension was reduced by averaging over all elements as we found them to be similar (Supplementary figure 12).

### GAMA computation algorithm

We summarize the computation of GAMA in Algorithm 1.

**Algorithm 1** Compute GAMA

```
procedure ComputeGAMA(sequences)
    function Gradients(LSTMmodel, sequences)
        for each seq_i ∈ sequences do
            for d1 ← 1, 22 do
                for d2 ← 1,lengh seq_i do
                    for d3 ← 1,lengh seq_i do
                        for d4 ← 1, 22 do
                            ig_tensor4d(d1, d2, d3, d4, i) ⟵ IG_{d1,d2}(LSTMmodel(seq)_{d3,d4})
                                                        ▷ for IG-function see Figure 10
                        end for
                    end for
                end for
            end for
        end for
        ig_tensor3d ⟵ meanDim3(ig_tensor4d)
        ig_tensor2d ⟵ pickOneHotEncodedElementsD2(ig_tensor3d)
        return ig_tensor2d
    end function
    initLSTMmodel ⟵ initilizeLSTM()
    trainedModel ⟵ trainLSTM(initLSTMmodel,sequences)
    initTensor ⟵ Gradients(initLSTMmodel, sequences)
    trainedTensor ⟵ Gradients(trainedModel, sequences)
    return GAMA(initTensor, trainedTensor)      ▷ for GAMA-function see Equation 1
end procedure
```

**Algorithm 1. Calculation of GAMA:** each sequence is passed into the trained LSTM model and the IG is computed for all outputs of the LSTM. This is done both for the reference model and the trained model. Subsequently, the difference is quantified using the GAMA function.

## Packages used

For the creation of graphics, we employed several libraries. Logo plots were generated using Logomaker 0.8 [89], while all other figures were created using Seaborn 0.11.1 [90].

Our computational work relied on the following libraries: PyTorch 1.8.1 [91], NumPy 1.23.1 [92], pandas 1.2.4 [93], and SciPy 1.14.1 [94]. These libraries were integral to the implementation and analysis of our methods.

# Funding

This was supported by a UiO: LifeScience Convergence Environment Immunolingo (to VG and GKS), a Norwegian Cancer Society Grant (#215817, to VG), Research Council of Norway projects (#300740, #331890 to VG), a Research Council of Norway IKTPLUSS project (#311341, to VG and GKS), Stiftelsen Kristian Gerhard Jebsen (K.G. Jebsen Coeliac Disease Research Centre) (to GKS) and the European Union (ERC, AB-AG-INTERACT, 101125630, to VG).

# Competing interests

V.G. declares advisory board positions in aiNET GmbH, Enpicom B.V, Absci, Omniscope, and Diagonal Therapeutics. V.G. is a consultant for Adaptive Biosystems, Specifica Inc, Roche/Genentech, immunai, LabGenius, and FairJourney Biologics. V.G. is an employee of Imprint LLC. PR declares employment by F. Hoffmann-La Roche AG.

# Supplementary Material

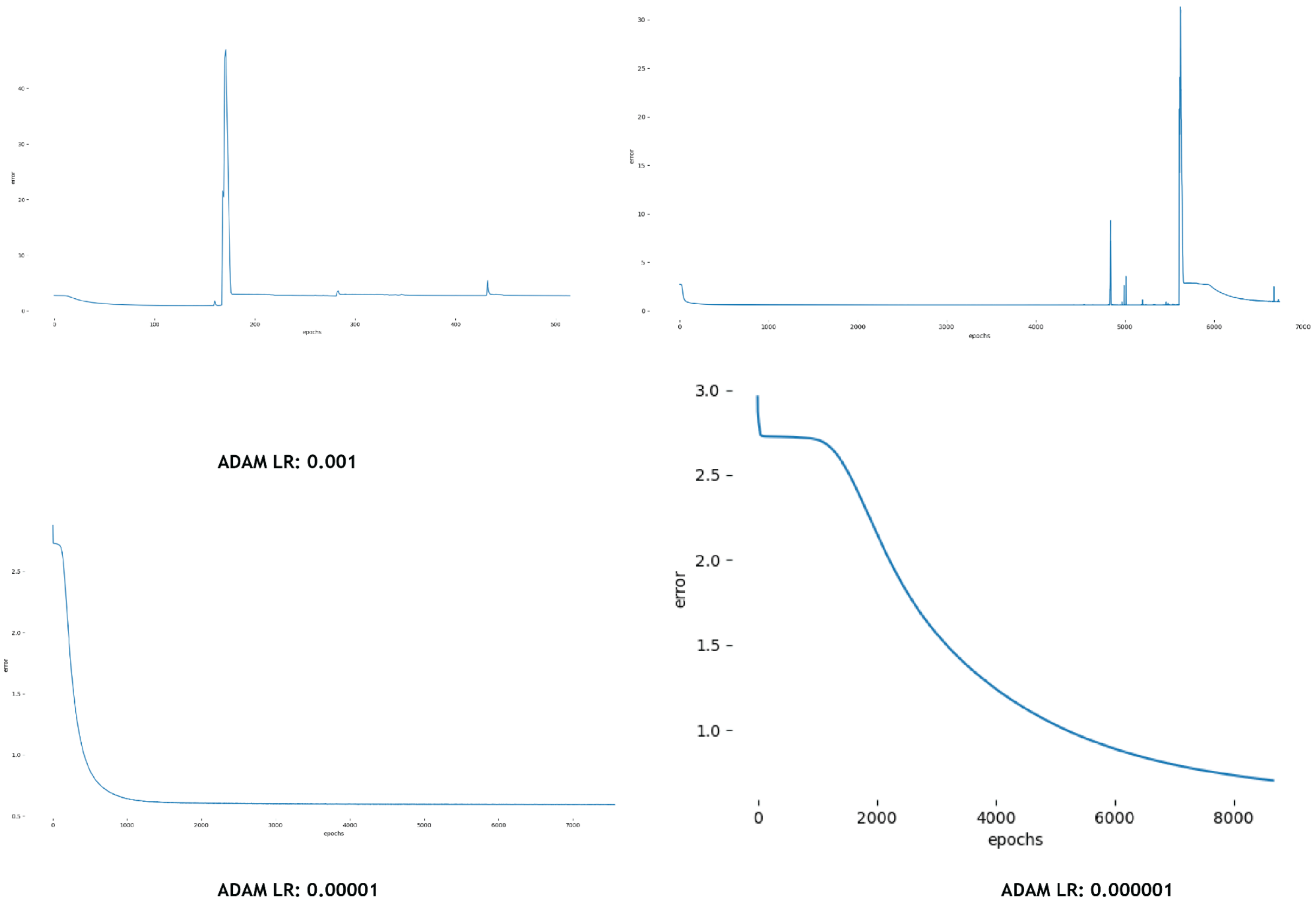

**Supplementary Figure 1 A learning rate of 0.00001 is effective for training the LSTM on the synthetic dataset.** Evaluating the influence of four different learning rates on the training stability on the synthetic dataset. The ADAM optimizer with a learning rate of 0.00001 showed a good trade-off between training speed and stability.

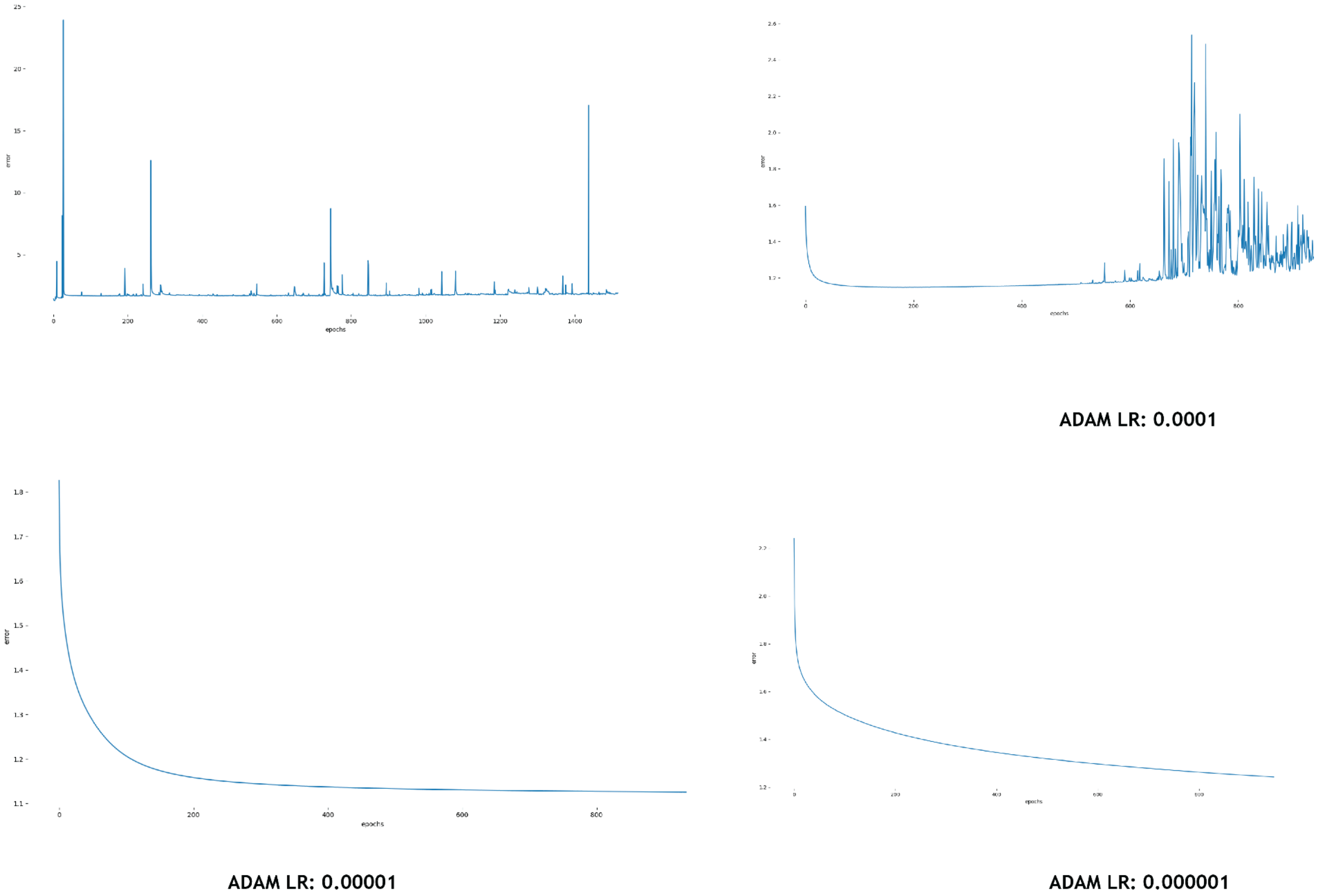

**Supplementary Figure 2 A learning rate of 0.00001 is effective for training the LSTM on the experimental dataset.** Evaluating the influence of four different learning rates on the training stability on the experimental dataset. The ADAM optimizer with a learning rate of 0.00001 showed a good trade-off between training speed and stability.

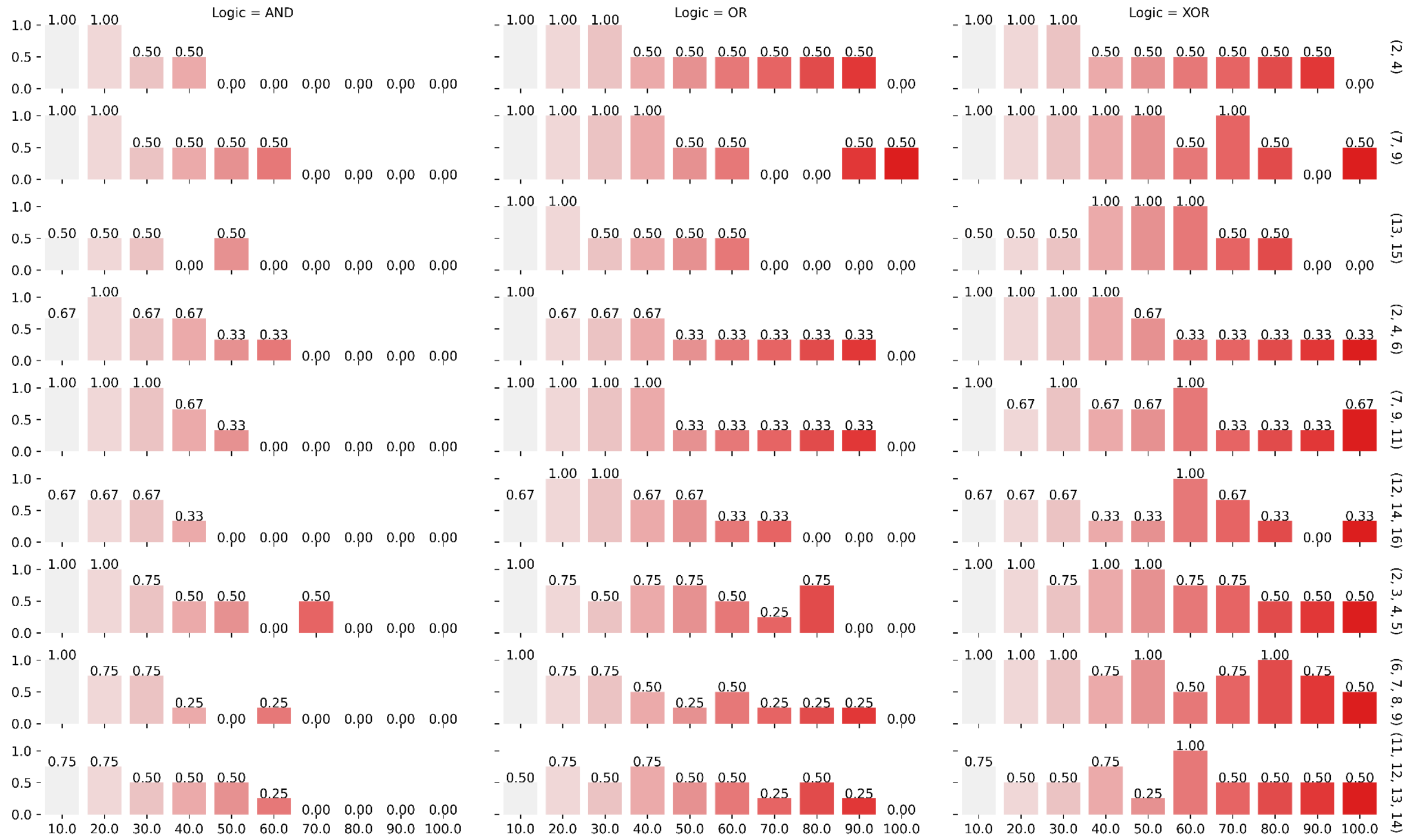

**Supplementary Figure 3 Enhanced retrieval efficiency for motifs positioned at the end of sequences.** The bar charts illustrate the retrieval effectiveness for motifs placed at the end of sequences. The secondary y-axis denotes the false-negative rate, while the secondary x-axis represents the signal-to-noise ratio. The primary x-axis categorizes the results based on three motif-logic variants: AND, OR, and XOR. The primary y-axis provides a comparison across different motif positions. Each bar in the chart corresponds to a unique dataset or experimental condition, encompassing a total of 270 bars. Thus, the figure comprehensively visualizes all synthetic sequence experiments, demonstrating that motif recovery rates are superior when motifs are located at the end of the sequence.

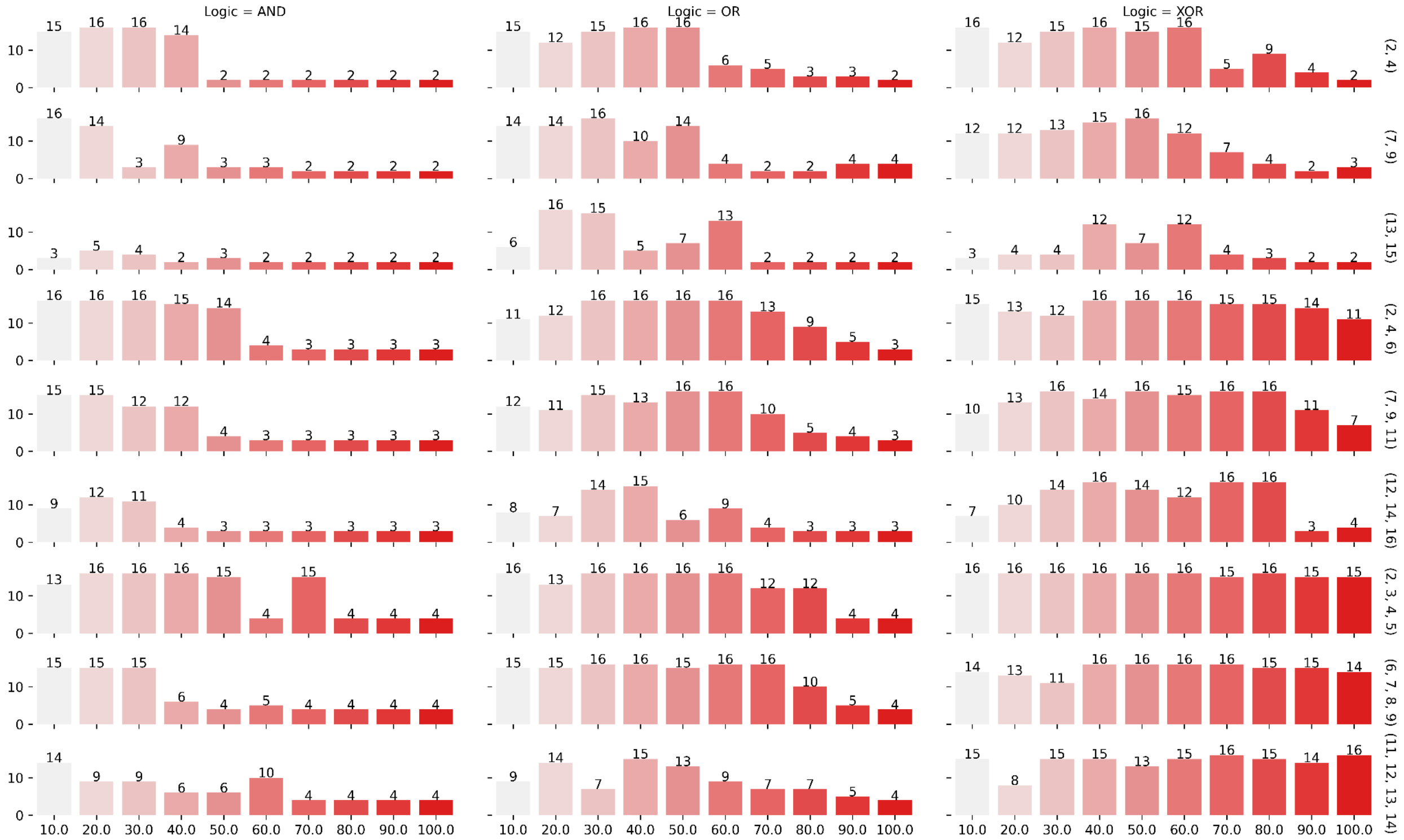

**Supplementary Figure 4 Enhanced retrieval efficiency for motifs positioned at the end of sequences.** The bar charts illustrate the top-k position metric, defined as the number of amino acid positions from the top of the list—as determined by GAMA—required to recover the entire motif, with smaller values indicating superior performance. The secondary y-axis displays the top-k metric, while the secondary x-axis represents the signal-to-noise ratio. The primary x-axis categorizes results based on three motif-logic variants: AND, OR, and XOR. The primary y-axis provides a comparison across different motif positions within sequences. Each bar corresponds to a unique dataset or experimental condition, totaling 270 bars. This figure comprehensively visualizes all synthetic sequence experiments, demonstrating that retrieval efficiency is significantly enhanced for motifs situated at the end of sequences.

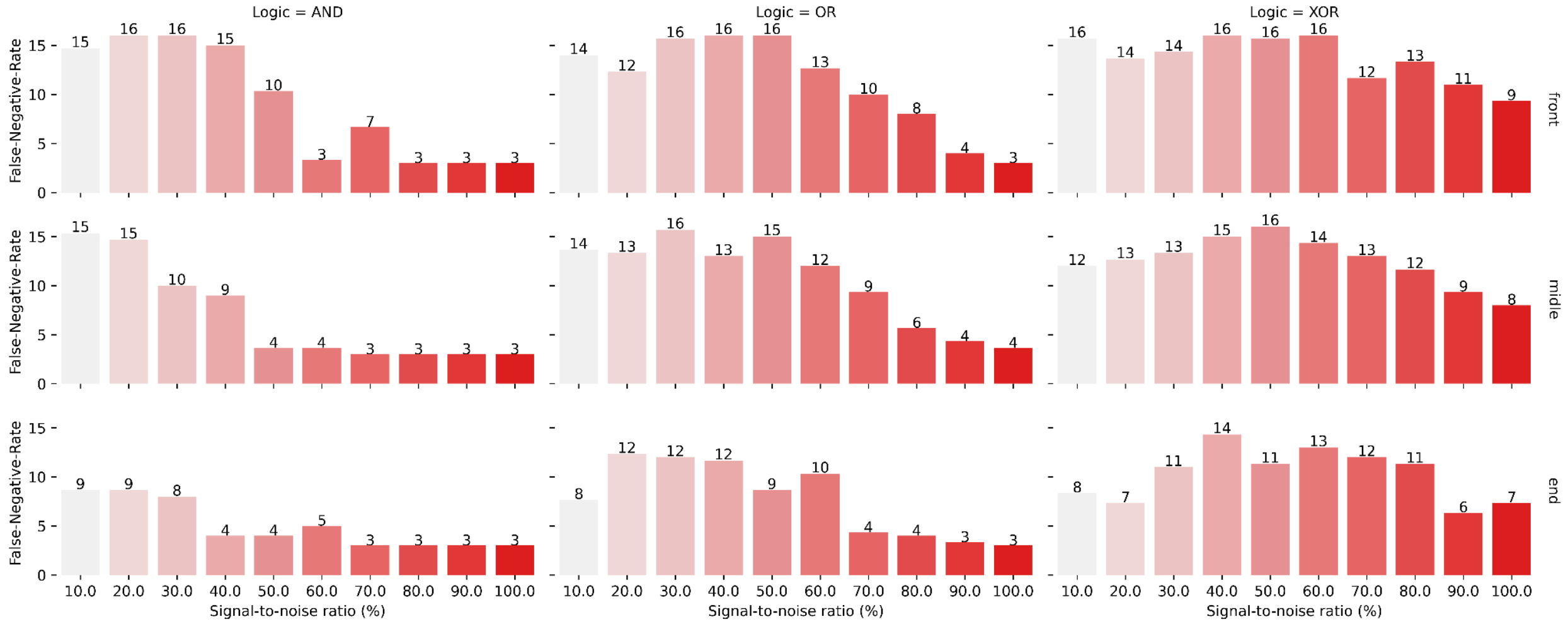

**Supplementary Figure 5 Enhanced retrieval efficiency for motifs positioned at the end of sequences.** The bar charts present the top-k position metric, which is defined as the count of amino acid positions from the top of the list—as determined by GAMA—until the entire motif is recovered, with smaller values indicating better performance. The secondary y-axis depicts the top-k metric, while the secondary x-axis represents the signal-to-noise ratio. The primary x-axis illustrates the results for three motif-logic variants: AND, OR, and XOR. The primary y-axis compares motifs located at the front, middle, and end of sequences, with values obtained by averaging different motif lengths. The figure indicates that motif recovery rates are superior when motifs are positioned at the end of the sequence.

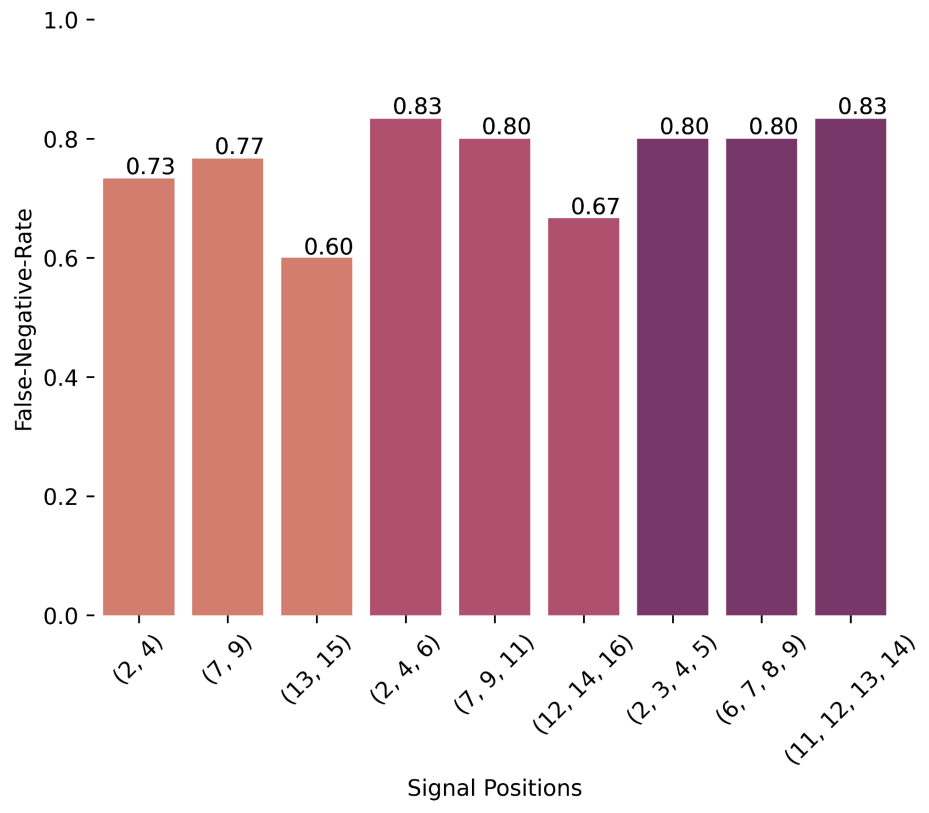

**Supplementary Figure 6: Retrieval performance improves for shorter motifs.** The y-axis indicates the false-negative-rate and the x-axis indicates the motif type and position. The three 2-position motifs have lower false-negative-rates than the longer (3-and-4)-motifs.

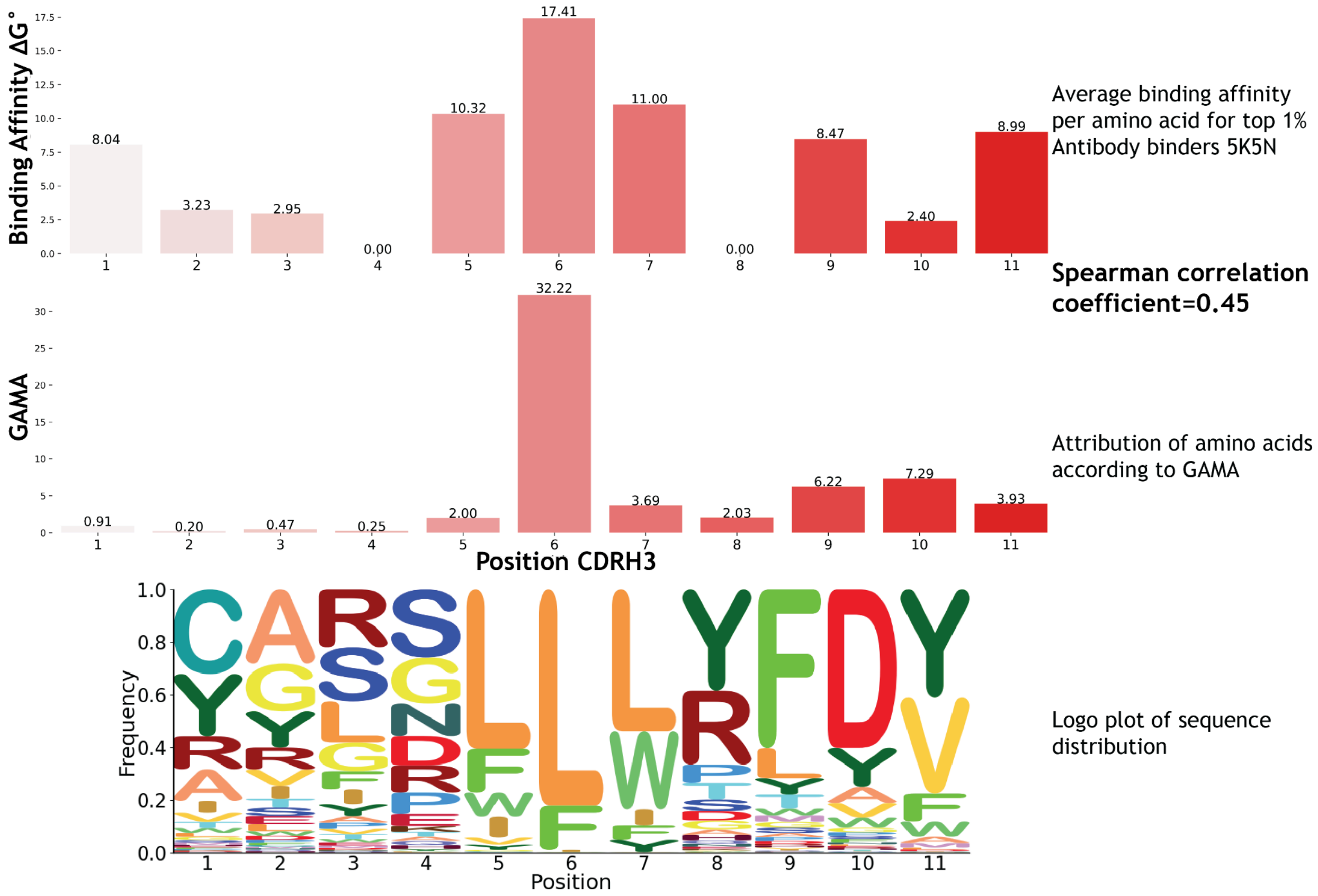

**Supplementary Figure 7 Absolut! Estimated amino acid vice binding affinities correlate with GAMA importance values.** Average amino acid binding affinities and of 2047 CDRH3 sequences of the top 1% binders to the 5K5N antibody and attribution assignment according to GAMA have a Spearman correlation of 0.45.

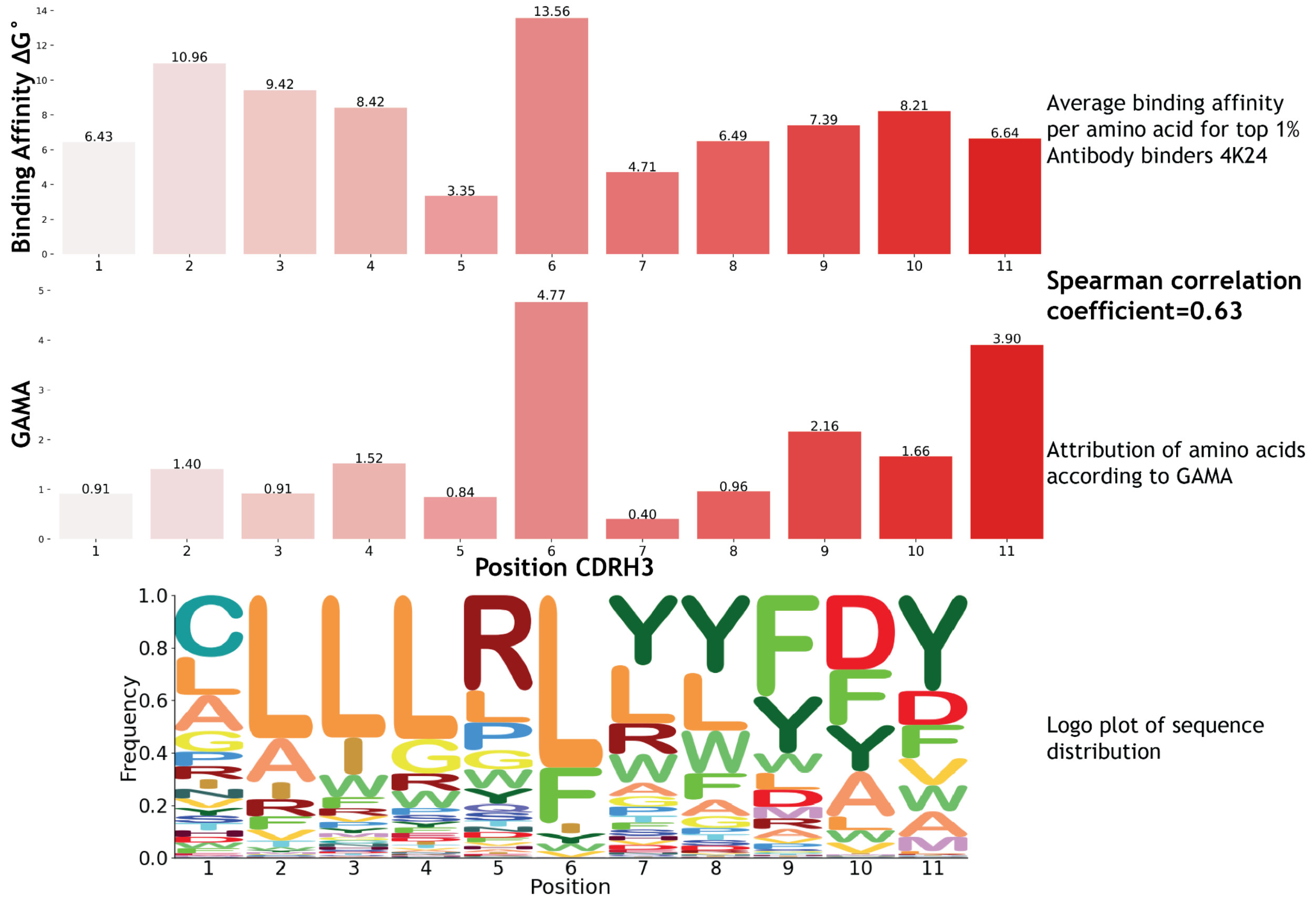

**Supplementary Figure 8 Absolut! Estimated amino acid vice binding affinities correlate with GAMA importance values.** Average amino acid binding affinities and of 2173 CDRH3 sequences of the top 1% binders to the 4K24 antibody and attribution assignment according to GAMA have a Spearman correlation of 0.63.

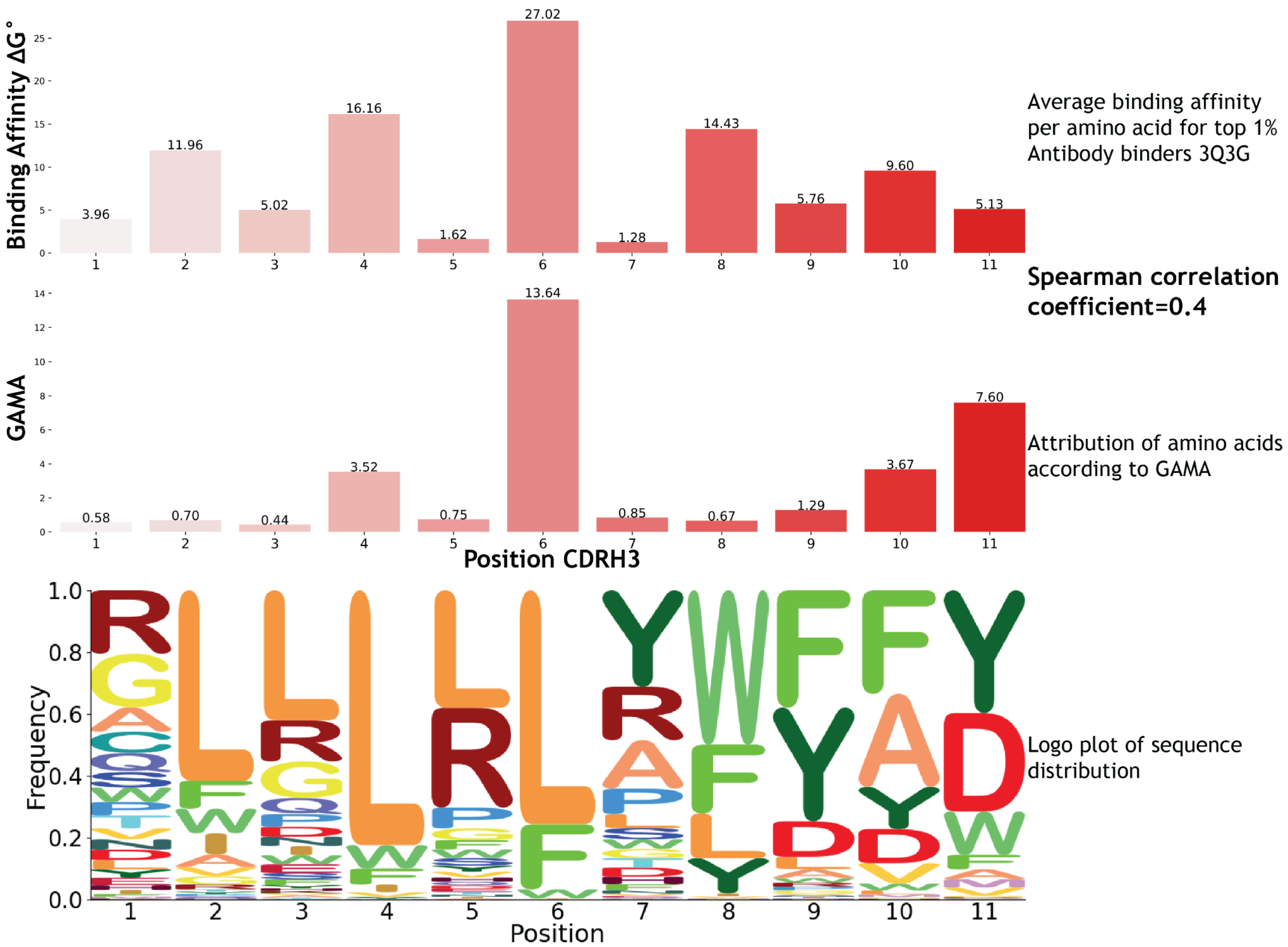

**Supplementary Figure 9 Absolut! Estimated amino acid vice binding affinities correlate with GAMA importance values.** Average amino acid binding affinities and of 1013 CDRH3 sequences of the top 1% binders to the 3Q3G antibody and attribution assignment according to GAMA have a Spearman correlation of 0.4.

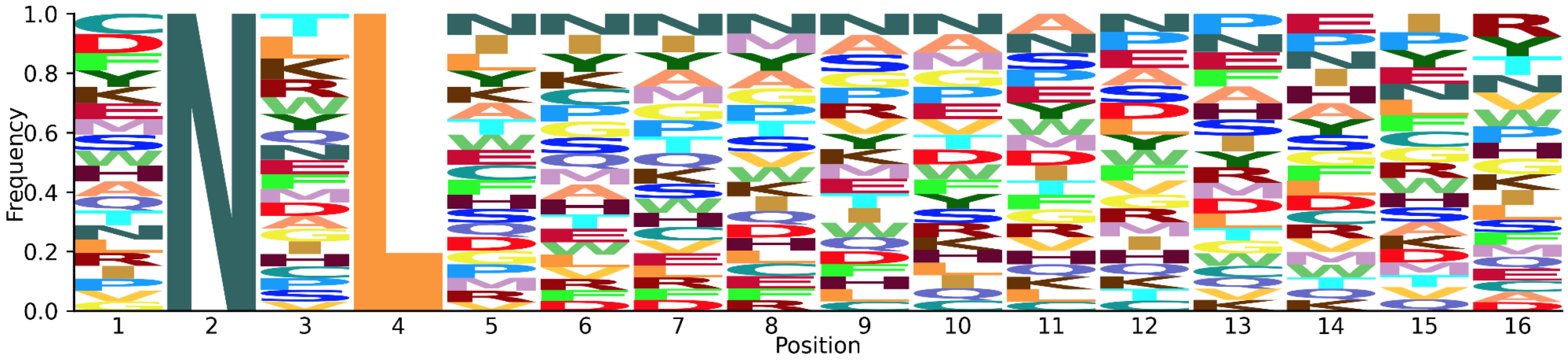

**Supplementary Figure 10 Generative LSTM models trained on the synthetic datasets produce representative samples of the training distribution.** Logoplot of the frequency distribution of 10 000 samples, sampled with a temperature parameter of 0.5, from a LSTM trained on a synthetic dataset with implanted motifs at positions two and four, "AND" logic interaction, and 100% signal-to-noise ratio.

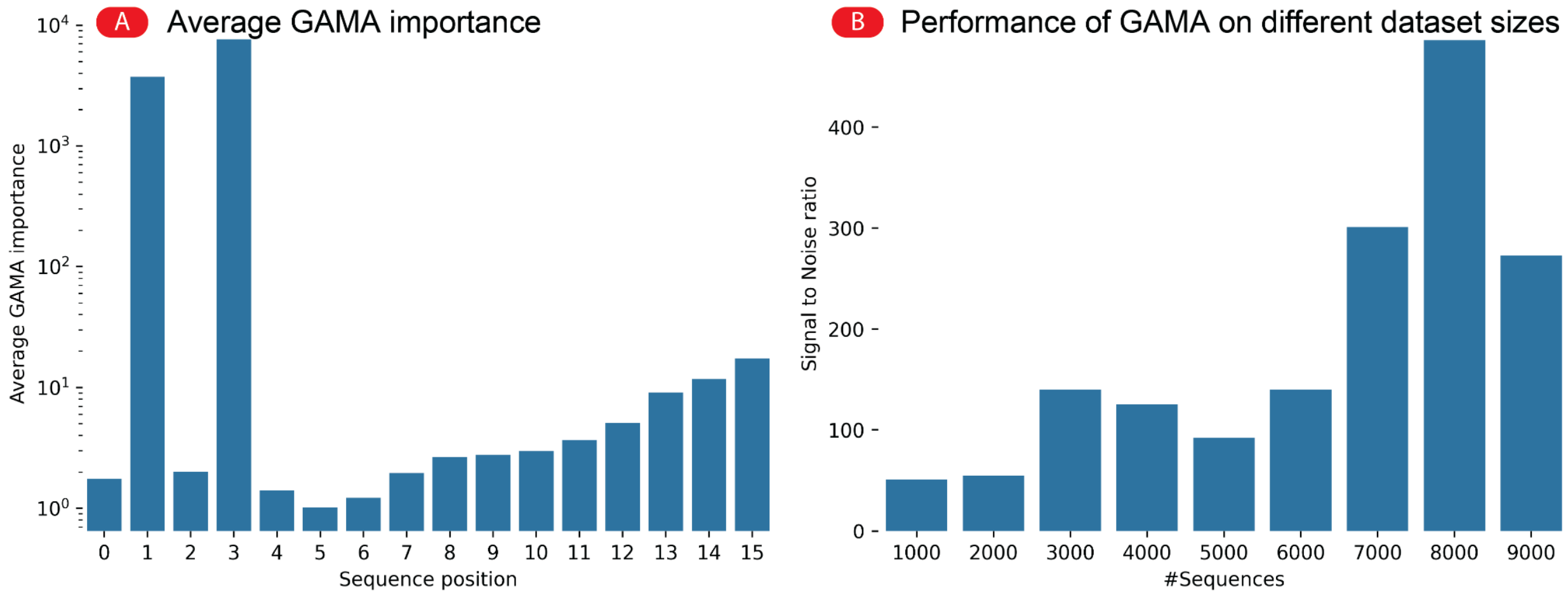

**Supplementary Figure 11. Larger number of sequences results in stronger GAMA signals.** We computed GAMA for nine synthetic datasets with a size between 1000 and 9000 sequences. All datasets had a signal implanted at positions two and four, motif positions interact with "AND" logic, and had no noise sequences. For all nine datasets, GAMA identifies the motif positions correctly. (A) This panel shows the average GAMA importance on a log scale, for each sequence position, over all nine datasets. A strong differentiation for the implanted motif positions can be seen at two and four.

(B) In this panel we investigated the ratio of GAMA importance values at positions where the motif was implanted (two and four) versus the GAMA at non-motif positions. Even datasets with as little as 1000 sequences, lead to signal to noise ratios far larger than one.

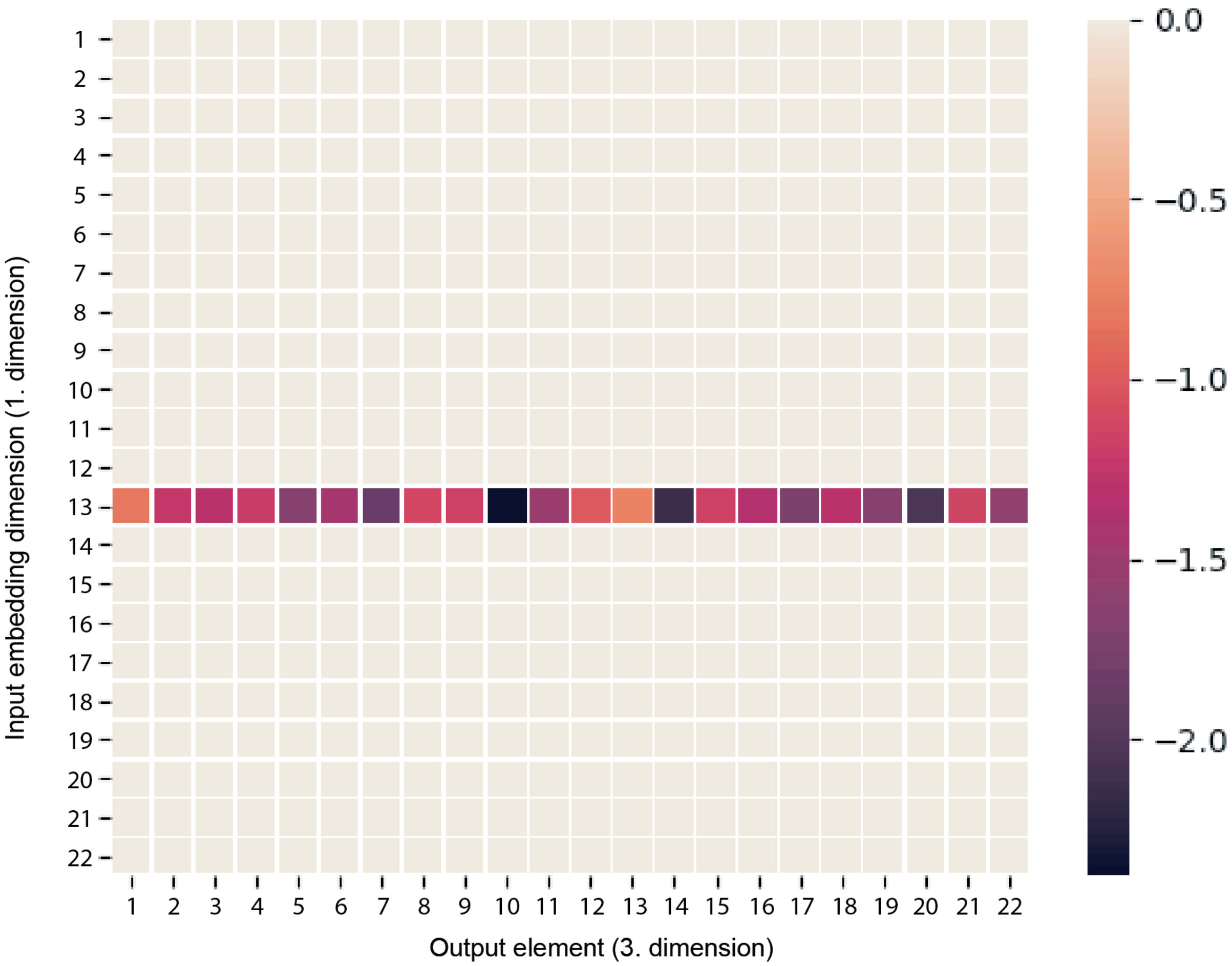

**Supplementary Figure 12 IG values computed for the 22 output dimensions are similar.** The heatmap values show IGs for a single sequence w.r.t. to one input token and computed for the next output token. The 22 encoding dimensions are displayed on the y-axis. The output dimension for which the gradient is computed, is displayed on the x-axis. The activation pattern is typical for all possible combinations of input-output combinations of the LSTM. The sequence displayed here had the input token 12 encoded as input. We can see that only IGs of row 12 are activated. The four dimensional IG tensor is reduced to a 2d tensor by selecting row 12 and averaging over it.